\documentclass{article}

\usepackage{arxiv}

\usepackage[utf8]{inputenc} 
\usepackage[T1]{fontenc}    
\usepackage{hyperref}       
\usepackage{booktabs}       
\usepackage{amsfonts}       
\usepackage{nicefrac}       
\usepackage{microtype}      
\usepackage{graphicx}
\usepackage{natbib}
\usepackage{doi}

\usepackage{microtype}

\usepackage{subfigure}
\usepackage[yyyymmdd]{datetime}

\usepackage[multiple]{footmisc}

\usepackage{mathtools}
\usepackage{fancyhdr}
\usepackage{multirow}
\usepackage{algorithm}
\usepackage{algorithmic}
\usepackage{physics}

\newcommand{\x}{\textbf{x}}
\newcommand{\p}{\textbf{p}}
\newcommand{\q}{\textbf{q}}

\newcommand{\Pb}{\textbf{P}}
\newcommand{\Q}{\textbf{Q}}

\newcommand{\X}{\textbf{X}}

\newcommand{\CC}{\textbf{C}}

\newcommand{\Rc}{\mathcal{R}}
\newcommand{\Normal}{\mathcal{N}}

\title{A Payload Optimization Method for Federated Recommender Systems}

\author{Farwa K. Khan \\
	Department of Computer Science, \\
	National University of Computer and Emerging Technology, \\
	Lahore, Pakistan. \\
	\texttt{L191881@lhr.nu.edu.pk} \\
	\And
	Adrian Flanagan\\
	Helsinki Research Center,\\
	Europe Cloud Service Competence Center,\\
	Huawei Technologies Oy (Finland) Co. Ltd,\\
	Helsinki, Finland\\
	\texttt{adrian.flanagan@huawei.com} \\
	\And
	Kuan E. Tan\\
	Helsinki Research Center,\\
	Europe Cloud Service Competence Center,\\
	Huawei Technologies Oy (Finland) Co. Ltd,\\
	Helsinki, Finland\\
	\texttt{kuan.eeik.tan@huawei.com} \\
	\And
	Zareen Alamgir\thanks{The authors contributed equally to this work.}\\
	Department of Computer Science, \\
	National University of Computer and Emerging Technology, \\
	Lahore, Pakistan.\\
	\texttt{zareen.alamgir\}@lhr.nu.edu.pk} \\
	\And
	Muhammad Ammad-ud-din\footnotemark[1]\\
	Helsinki Research Center,\\
	Europe Cloud Service Competence Center,\\
	Huawei Technologies Oy (Finland) Co. Ltd, \&\\
	ComparablesAI Oy, Helsinki, Finland\\
	\texttt{mu.ammad.ud.din@comparables.ai} \\
}

\date{July 10, 2021}	

\renewcommand{\headeright}{A preprint - \date{July 10, 2021}}
\renewcommand{\shorttitle}{FL payload optimization}

\hypersetup{
pdftitle={A Payload Optimization Method for Federated Recommender Systems},
pdfauthor={Farwa K. Khan, Adrian Flanagan,Kuan E. Tan,Zareen Alamgir,Muhammad Ammad-ud-din},
pdfkeywords={collaborative filtering, federated learning, multi-arm bandits, payload optimization, recommender systems},
}

\begin{document}
\maketitle

\begin{abstract}
	In this study, we introduce the payload optimization method for federated recommender systems (FRS). In federated learning (FL), the global model payload that is moved between the server and users depends on the number of items to recommend. The model payload grows when there is an increasing number of items. This becomes challenging for an FRS if it is running in production mode. To tackle the payload challenge, we formulated a multi-arm bandit solution that selected part of the global model and transmitted it to all users. The selection process was guided by a novel reward function suitable for FL systems. So far as we are aware, this is the first optimization method that seeks to address item dependent payloads. The method was evaluated using three benchmark recommendation datasets. The empirical validation confirmed that the proposed method outperforms the simpler methods that do not benefit from the bandits for the purpose of item selection. In addition, we have demonstrated the usefulness of our proposed method by rigorously evaluating the effects of a payload reduction on the recommendation performance degradation. Our method achieved up to a 90\% reduction in model payload, yielding only a $\sim$4\% - 8\% loss in the recommendation performance for highly sparse datasets\footnote[1]{Accepted for publication in Fifteenth ACM Conference on Recommender Systems (RecSys ’21), September 27-October 1, 2021, Amsterdam, Netherlands. ACM, New York, NY, USA, 16 pages. \url{https://doi.org/10.1145/3460231.3474257}}. 
\end{abstract}

\keywords{collaborative filtering \and federated learning \and multi-arm bandits \and payload optimization \and recommender systems}

\section{Introduction}\label{sec:introduction}

Federated Learning (FL)~\cite{mcmahan2017communication}, a privacy-by-design machine learning approach, has introduced new ways to build recommender systems (RS). Unlike traditional approaches, the FL approach means that there is no longer a need to collect and store the users’ private data on central servers, while making it possible to train robust recommendation models. In practice, FL distributes the model training process to the users’ devices (i.e., the client or edge devices), thus allowing a global model to be trained using the user-specific local models. Each user updates the global model locally using their personal data and sends the local model updates to a server that aggregates them according to a pre-defined scheme. This is in order to update the global model.

A prominent direction of research in this domain is based on Federated Collaborative Filtering (FCF)~\cite{ammad2019federated,chai2019secure,dolui2019poster} that extends the standard Collaborative Filtering (CF)~\cite{Hu2008} model to the federated mode. CF is one of the most frequently used matrix factorization models used to generate personalized recommendations either independently or in combination with other types of model~\cite{koren2009matrix}.  Essentially, the CF model decomposes the user-item interaction (or rating) data into two sets of low-dimensional latent factors, namely the user-factor and item-factor, therefore capturing the user and item specific dependencies from the interaction data respectively. The learned factors are then used to generate personalized recommendations regarding the items that the users have not interacted with before.

The FCF distributes parts of the model computation so then all of the item-factors (i.e., the global model) are updated on the FL server and then distributed to each user. The user specific factors are updated independently and locally on each device using the user’s private data and the item-factors received from the server. The local model updates through the gradients are then calculated for all of the items on each user’s device. This is then transmitted to the server where the updates are aggregated to update the item-factors (also known as the update of the global model). To achieve model convergence, FCF and similar federated recommendation models require several communication rounds (of global vs. local model updates) between the FL server and the users. In each round, the computational payload (also known as the carrying capacity of a packet or transmission data unit) that is transferred (upload/download) across the network and between the server and users depends on the size of the global model (here it is the $\text{number of items} \times \text{number of factors}$).    

Beyond the major challenges of FL systems~\cite{litian2019federated,li2019federated}, there exists a practical concern that arises when running large-scale federated recommender systems (FRS) in production.  Considering the number of factors to be fixed, the model payload increases linearly with the increase in the number of items. Table ~\ref{Payload} demonstrates the expected payload estimations of a global model with a total number of items between 3000 -- 10 million. For a large-scale FRS comprised of 100,000 items, there exists a key problem of an increasing payload not only for the users but also for the broadband/mobile internet service providers and operators. The requirement to transmit huge payloads between the FL server and users over several communication rounds imposes strict limitations for a real-world large-scale FL based recommender system. 

\begin{figure}
	\centering
	\includegraphics[width=0.60\textwidth]{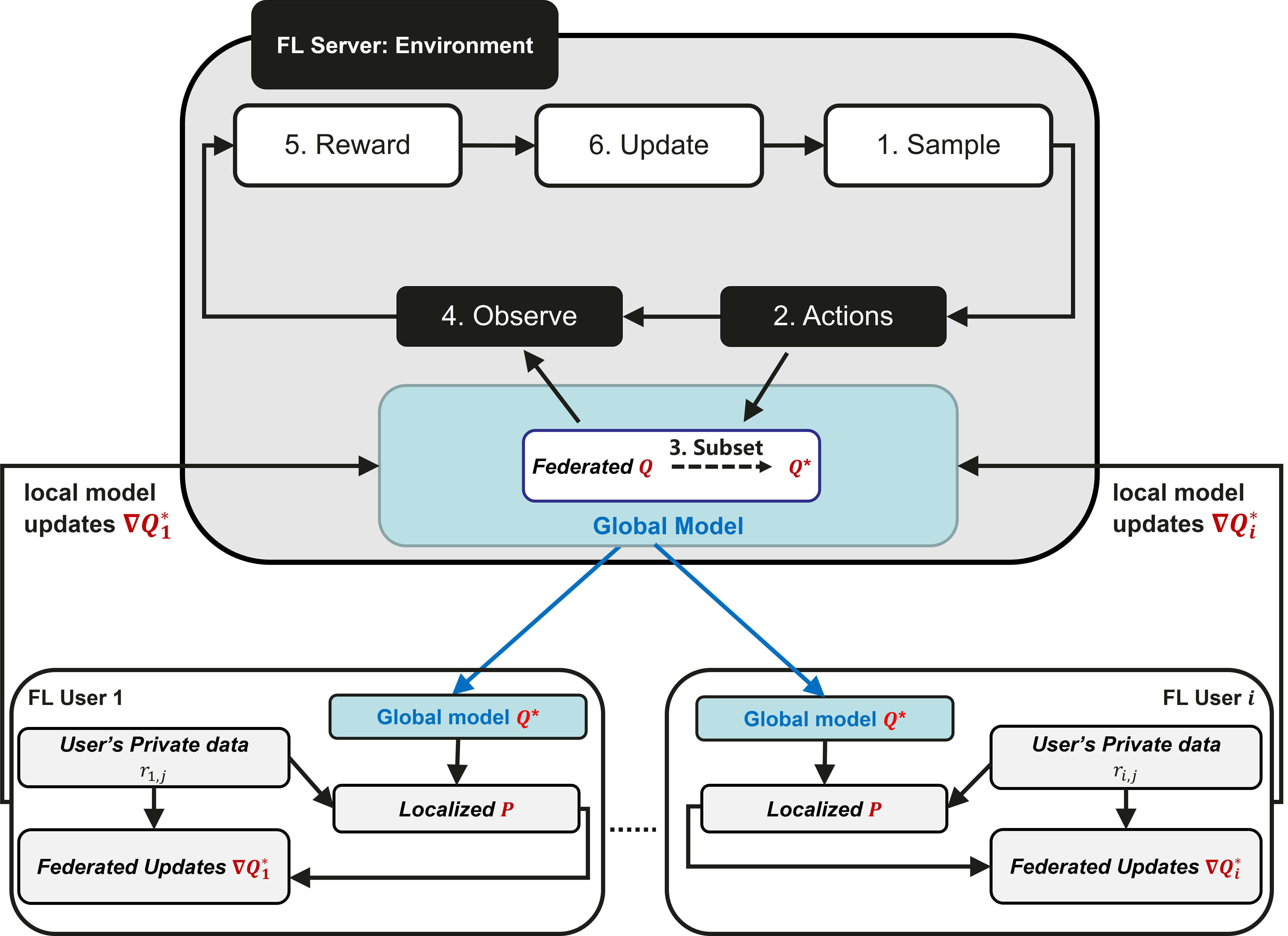}
	\caption{The payload optimization method proposed in this study for federated recommender systems (FRS). The payload optimization is performed on the FL server using a multi-arm bandit solution. 1. The bandit model samples a set of items from a probability distribution. 2. The bandit takes a certain action by selecting a particular set of items. 3. The FL server selects part of the global model $\Q$ based on the item suggested by the bandit model. The selected $\Q^*$ is transmitted to all users. The updates through the gradients of $\nabla \Q^*$ are computed on each user’s device and transmitted to the FL server to update $\Q$. 4. The local model updates of $\nabla \Q^*$ are considered to be feedback. 5 The rewards are estimated for the selected items based on their feedback. 6. The estimated rewards are used by the bandit model to update the parameters of the probability distribution.}
	\label{flow_diagram}
\end{figure}

\begin{table}[htbp]
	\centering
	\caption{The Federated Collaborative Filtering model’s payload increases linearly with the increasing number of items. For a large-scale FL recommender system comprised of millions of items, the payload can exceed 1GB. Assuming a fixed number of factors = 20, below the payloads were estimated assuming a floating point precision of 64 and 8 bits per 1 byte. The simple formula used to estimate the payload is (\#parameters $\times$ 64) / 8 = Bytes. The FL model training with increased payloads becomes challenging for resource-constrained devices as well as for network operators with a limited communication bandwidth.} 
	\begin{tabular}{c c c c c c c }
		\hline \hline 
		\textbf{\# Items} & 3912 & 10k & 100k & 500K & 1 M & 10 M \\
		\textbf{Payload (approx)} & 625KB & 1.6 MB  & 16 MB & 80 MB  & 160 MB & 1.6 GB  \\
		\hline
	\end{tabular}%
	\label{Payload}%
\end{table}%
To tackle the payload challenge, we present a new payload optimization method for FRS as shown in Figure~\ref{flow_diagram}.  We adopted multi-armed bandit (MAB), a classical approach to reinforcement learning, in order to formulate our solution for minimizing the payloads. In each communication round, our optimization method intelligently selects part of the global model to be transmitted to all users. The selection process is guided by a bandit model with a novel reward policy well-suited for FRS. In this way, instead of transmitting (uploading/downloading) the huge payload that includes the entire global model, only part of the global model with a smaller payload is transmitted over the FL network. The users perform the standard model updates as part of the FRS~\cite{ammad2019federated,10.1007/978-3-030-67661-2_20},  thus avoiding any additional optimization steps (see Figure~\ref{flow_diagram}). As a case study, we have presented the payload optimization of a traditional FCF method. However, the proposed method can be generalized to advanced deep learning-based FL recommendation systems~\cite{qi2020privacy} and it can also be applied to a generic class of matrix factorization models~\cite{10.1007/978-3-030-67661-2_20}. We extensively compared the results from three benchmark recommendation datasets, specifically Movielens, Last-FM, and MIND. The findings confirm that the proposed method consistently performed better than the baseline method and achieved a 90\% reduction in payload with an average recommendation performance degradation ranging from $\sim$4\% to 8\% for highly sparse datasets (Last-FM and MIND).

The contribution of this work is two-fold: (1) We have proposed the first method to optimize the payload in FRS and (2) We have empirically demonstrated the usefulness of our proposed method by rigorously evaluating the effects of payload reduction on recommendation performance.

\section{Methods}\label{sec:method}

\subsection{Collaborative Filtering (CF)}\label{subsec:cf_method}
Given a collection of user-item interactions $\x_{i} \in \Rc^M$ for $i = 1 \ ,\dots, \ N$ users and $M$ items collected in a data matrix $\X \in \Rc^{N \times M}$ , the standard CF~\cite{koren2009matrix} is defined as a matrix factorization model: 
\begin{equation}
	{x}_{ij} \sim \p_i^T \q_j
	\label{eq:mf}
\end{equation}
The CF model factorizes ${x}_{ij}$ into a linear combination of low-dimensional latent item-factors $\q_j \in \Rc^K$ for $j =1 \ ,\dots, \  M$ and user-factors $\p_i \in \Rc^K$ for $i = 1 \ ,\dots, \ N$ collected in factor matrices $\Q \in \Rc^{K \times M}$ and $\Pb \in \Rc^{K \times N}$ respectively, where $K$ is the number of factor. The cost function optimizing across all users and items is then given as:
\begin{equation}
	J =  \sum_i \sum_j c_{ij} (x_{ij} - \p_i^T \q_j)^2 + \lambda \Big(\sum_i \lVert \p_i \rVert^2 + \sum_j \lVert \q_j\rVert^2\Big)
	\label{eq:opt}
\end{equation}
where a confidence parameter $c_{ij} = 1 + \alpha x_{ij};\ \alpha >0$  is introduced to account for the uncertainties arising from the unspecified interpretations of $x_{ij}$ in implicit feedback scenario. Specifically, $x_{ij}\geq1$ denotes that the user $i$ has interacted with the item $j$. However, $x_{ij}=0$ can have multiple interpretations such as the user $i$ does not like the item $j$ or maybe the user $i$ is oblivious to the existence of the $j$th item~\cite{Hu2008}. Lastly, $\lambda$ is the L2-regularization parameter set to avoid over-fitting.  

\subsection{Federated Collaborative Filtering (FCF)}\label{subsec:fcf_method}
FCF extends the classical CF model to the federated mode~\cite{ammad2019federated,chai2019secure,dolui2019poster}. FCF distributes parts of the model computation (Eq.~\ref{eq:opt}) to the user’s device as illustrated in Figure~\ref{flow_diagram}. The key idea is to perform local training on the device so then the user’s private interaction data (e.g., ratings or clicks) is never transferred to the central server. The global model is updated on the server after the local model updates have been received from a certain number of users. Specifically, for a particular user $i$, the federated update of the private user-factor $\p_i$ is performed independently without requiring any other user’s private data. The optimal solution is obtained by taking $\partial J(\p_i^*)/\partial \p_i = 0$, setting $\p_i = \p_{i^*}$, from Eq.~\ref{eq:opt}
\begin{equation}
	\p_i^* \ = \ \big(\Q\CC^i\Q^T + \lambda I\big)^{-1}\Q\CC^i \x_i,
	\label{eq:x_opt}
\end{equation}
Importantly, the update depends on the item-factor $\Q$ which is received from the FL server for each round of model updates. However, the item-factor $\Q$ is updated on the FL server using a stochastic gradient descent approach.
\begin{equation}
	\Q \ = \ \Q  \ - \ \eta \sum_{i=1}^{\Theta} \nabla \Q_{i}
	\label{eq:gradientDescent}
\end{equation}
for some gain parameter $\eta$ and number of federated model updates $\Theta$ to be determined. A particular user $i$ computes the item gradients $\nabla \Q_{i}$  independently of all other users as 

\begin{equation}
	\nabla \Q_{i} = \left[\begin{array}{ccc}
		\frac{\partial J_i}{\partial q_{1,1}} & \cdots & \frac{\partial J_i}{\partial q_{1,M}} \\
		\vdots & \ddots & \vdots \\
		\frac{\partial J_i}{\partial q_{K,1}} & \cdots & \frac{\partial J_i}{\partial q_{K,M}}
	\end{array}\right]
\end{equation}

where $\frac{\partial J_i}{\partial \q_{j}} = \left[ \frac{\partial J_i}{\partial q_{1,j}} \cdots  \frac{\partial J_i}{\partial q_{K,j}} \right]$ for item $j$ is defined as:
\begin{equation}
	\frac{\partial J_i}{\partial \q_j} \ = \ -2 \ \big[c_{ij} (x_{ij} - \p_i^T\q_j)\big] \p_i  + 2 \lambda \q_j.
	\label{eq:sumf}
\end{equation}

\noindent Each user transmits the $\nabla \Q$ gradients of all items as local model updates to the FL server, where the $\nabla \Q$s are aggregated to update the global model $\Q$ (see Eq.~\ref{eq:gradientDescent}). The Adaptive Moment Estimation (Adam) ~\cite{kingma2015adam} method is used in the context of FCF~\cite{ammad2019federated,10.1007/978-3-030-67661-2_20} to better adapt the learning rate ($\eta$) to support faster convergence and greater stability. Finally, in order to compute the recommendations ${x_i}^* = \p_i^T \Q$, the user $i$ downloads the global model from the FL server according to a predefined configuration setting.

Importantly, in each FL training iteration, the model payloads $\Q$ and $\nabla \Q$ are transferred between the server and users, and vice-versa. The payload scales linearly with the increasing number of items (as shown in Table~\ref{Payload}). We next present our method for optimizing the model payloads by reducing the size of $\Q$ and $\nabla \Q$ to the point where it is suitable for FRS when deployed in production. 

\section{Payload Optimization method for Federated Collaborative Filtering}\label{subsec:payload_method}
We formulate a multi-arm bandit method to optimize $\Q$ model payloads for federated recommender systems (FRS). There exist numerous challenges when optimizing payloads. First, the FCF server does not know the user’s identity. Each user sends the $\nabla \Q$ updates which are aggregated without referencing any one user’s identity. To optimize the payload, we cannot determine the item memberships in terms of groups of users, therefore potentially relevant items may be selected in the $\Q$ model. Second, in contrast to the standard (offline) training of models, the FL training is performed online with the federated updates $\nabla \Q$ arriving from users in a continuously asynchronous fashion. In each iteration, $\Q$ is updated when the number of collected $\nabla \Q$ updates reaches a certain threshold $\Theta$. Several factors make the FL training computationally challenging such as a low number of users participating in the update, a lesser frequency of updates being sent by the users, and most importantly, a lousy communication over the Internet and the related network latency. 
In practice, the FCF model training is a complex online sequential learning problem that motivated our choice of proposed method for payload optimization. Consider a particular FCF-based recommendation model training set-up where at each FL iteration $t$, 
\begin{enumerate}
	\item the FL server requests the set of items (potential \textit{--arms}) from the bandit model,
	\item the bandit model selects a subset of items $M_s$ among the set of available $M$ items,
	\item the FL server only transmits the global model $\Q^*$ comprised of the selected items to $N$ users (or clients),
	\item a user $i$ for $i = 1 \ ,\dots, \ N$ returns feedback $s^{j}_t=\nabla^j \Q_{t}^*$ for $j = 1 \ ,\dots, \ M_s$ as the gradients of the selected items. 
\end{enumerate}
In our context, the feedback $s_t$ is used to compute the quantity that has to be optimized aka. reward $r_t$.  To handle the online sequential aspect of the FL model training, our bandit solution is composed of two main ingredients: (1) a strategy recommending the items in order to select the optimal $\Q^*$, and (2) a function to infer the rewards when using the feedback received from the FL users.  We refer to the proposed method as FCF-BTS (throughout the manuscript) and outline the FCF-BTS algorithmic steps in Algorithm~\ref{alg:fl_payload}.
\begin{algorithm}[t]
	\small
	\caption{FCF-BTS: Payload optimization for Federated Collaborative Filtering}
	\label{alg:fl_payload}
	\begin{algorithmic}[1]
		\STATE \textbf{FL Server}
		\STATE Set number of items to sample $M_s$
		\STATE Initialize global model $\Q$ matrix and update threshold $\Theta$
		\STATE Initialize Bayesian Thompson Sampling bandit model $BTS$
		\STATE Initialize local model updates $\nabla^j \Q = 0 \ \forall \ j = 1, 2 ...M$ matrix
		\STATE Initialize parameter to record exponential decay of the squared gradient $\upsilon^j = 0 \ \forall \ j = 1, 2 ...M$ 
		\FOR {$t = 1, 2 ...$}
		\STATE Select $M_s$ items from $BTS$ representing the largest sampled values ordered by their expected rewards \textbf{Eqs. \ref{eq:posterior},  \ref{eq:likilhood}}
		\STATE Subset the $\Q$ factor matrix based on $M_s$ items, denoted as ${\Q_{t}^*}$
		\STATE Transmit $\Q_{t}^*$ $\rightarrow$ \textbf{FL users}
		\STATE Receive item-factor gradients: ${\ \forall \ j = 1, 2 ...M_s; \  {\nabla^j} \Q_{t}^*} $ $\leftarrow$ \textbf{FL users}
		\IF{$\mbox{NumberGradientUpdates}>=\mbox{$\Theta$}$}
		\STATE Update $\Q$ based on ${\nabla} \Q_{t}^*$ \textbf{Eq.\ref{eq:gradientDescent}}
		\STATE Update $\upsilon_t^j$ using $\nabla^j \Q_{t}^{*}$ \textbf{Eq.\ref{eq:adam_v}}
		\FOR {$j = 1, 2 ...M_s$}
		\STATE Compute reward $r^j_t$ using \textbf{Eq.\ref{eq:reward}}
		\STATE Update $BTS$ parameters using $r^j_t$  \textbf{Eqs.\ref{eq:estimated_reward}, \ref{eq:posterior_param_variance}, \ref{eq:posterior_param_mean}}
		\STATE Update $\nabla^j\Q = \nabla^j\Q_{t}^{*}$ 
		\ENDFOR
		\ENDIF
		\ENDFOR
	\end{algorithmic}
\end{algorithm}

Formally, our bandit method for payload optimization is a tuple consisting of four elements  $<M_s,S,A,R>$:\\
\textbf{Item} $M_s$ is a subset of the items among the set of available $M$ items.\\
\textbf{State} $S = [S^1, S^2 \ ,\dots, \ S^{M_s}]$ is the set containing the feedback (or observations) collected by the bandit model from the FL environment. Particularly, $S^j = [s^{j}_1 \ ,\dots, \ s^{j}_t]$, where $s^{j}_t$ includes the feedback that the item $j$ (for $j = 1 \ ,\dots, \ M_s$ ) has received from the FL users at the iteration $t$. We consider $s^j_t$ to be the feedback that contains the local model updates $\nabla \Q^*$.\\
\textbf{Actions} $A = [A^1, A^2 \ ,\dots, \ A^{M}]$ is the set including the actions suggested by the bandit model. Specifically, $A^j = [a^{j}_1 \ ,\dots, \ a^{j}_t]$, where $a^{j}_t$ denotes the action taken by the bandit to recommend the item $j$ (for $j = 1 \ ,\dots, \ M$ ), to be included in $\Q^*$ at FL iteration $t$ .\\ 
\textbf{Reward} $R = [R^1, R^2 \ ,\dots, \ R^{M_s}]$, where $R: S \times A \rightarrow \Rc$ is the reward function. Particularly, $R^j = [r^{j}_1 \ ,\dots, \ r^{j}_t]$ where $r^{j}_t$ represents the reward for item $j$ (for $j = 1 \ ,\dots, \ M_s$ ) in each FL iteration $t$. After an action $a^j_t$ is taken by the bandit model, the user provides feedback $s^j_t$,  which is then used to estimate the reward using Eq.~\ref{eq:reward}. 
\subsection{Sampling Strategy}
As an item-based payload selection strategy, we used the widely known Bayesian Thompson Sampling (BTS)~\cite{thompson1933likelihood,thompson1935theory,chapelle2011empirical,scott2010modern,kawale2015efficient} approach with Gaussian priors for the rewards. We formulated a probabilistic model to sample the next set of item from the posterior distributions, which were then used for selecting $\Q^*$ optimally. Specifically, we assumed that the model of item rewards followed a normal distribution with an unknown mean and fixed precision ($ \tau = \frac{1}{variance}=1$) as given by:
\begin{equation}
	p(R^j|\mu^j,1) \sim \Normal(\mu^j, 1)
	\label{eq:likilhood}
\end{equation}
The prior probability for unknown $\mu^j$ for an item $j$ is also believed to be normally distributed with parameter $\mu_\theta$ and precision $\tau_\theta$ such as: 
\begin{equation}	
	p(\mu^j) \sim \Normal(\mu_\theta, \frac{1}{\tau_\theta})
	\label{eq:prior_mu}
\end{equation}
The posterior probability distribution of the unknown $\mu^j$ was obtained by solving the famous Bayes theorem~\cite{gelman2013bayesian}:
\begin{equation}
	p(\mu^j|R^j,\mu_\theta,\tau_\theta) \sim \Normal(\hat{\mu}^j_{\theta}, \frac{1}{\hat{\tau}^j_{\theta}})
	\label{eq:posterior}
\end{equation}
where the updates for the posterior parameters of the prior $\mu^j$ are estimated as~\cite{fink1997compendium,gelman2013bayesian}:
\begin{equation}
	\hat{\mu}^j_{\theta} = \frac{\tau_\theta \mu_\theta + n^j Z_t(a^j_t)}{\tau_\theta  + n^j}
	\label{eq:posterior_param_mean}
\end{equation}
where $n^j$ is the number of times that the item $j$ has been selected as part of $\Q^*$.  
\begin{equation}
	\hat{\tau}^j_{\theta} = \tau_\theta + n^j \tau
	\label{eq:posterior_param_variance}
\end{equation}
In Eq. \ref{eq:posterior_param_mean}, $Z_t(a^j)$ is the estimated value of action $a^j$ at FL iteration (or time step) $t$ and given by:
\begin{equation}
	Z_t(a^j_t) = \frac{1}{n^j} \sum_{i=1}^{n^j} r^j_t
	\label{eq:estimated_reward}
\end{equation}
where $r^j_t$ (Eq.~\ref{eq:reward}) is the reward obtained at FL iteration $t$ when action $a^j_t$ was taken. 
Essentially, in each FL iteration $t$, we update two parameters $\hat{\mu}^j_{\theta}$ and $\hat{\tau}^j_{\theta}$ of the selected item $j \in 1 \ ,\dots, \ M_s$.  Next, we sampled $\mu^j$ from the posterior distribution (specified in Eq.~\ref{eq:posterior}) before selecting the items (aka. \textit{--arms}) corresponding to the largest sampled values ordered by their expected rewards (Eq.~\ref{eq:likilhood}). Our setting is similar to that of the multiple arms selection ($top-M$) problem in RS~\cite{streeter2008online,radlinski2008learning,uchiya2010algorithms,louedec2015multiple}, where numerous studies have concluded that BTS achieved a substantial reduction in running time compared to non-Bayesian simpler sampling strategies~\cite{gopalan2014thompson,broden2018ensemble}. 
\subsection{Reward Function}
In this section we present a novel reward function designed for FRS. At the FL iteration $t$, the sampling strategy recommends item set $ \{1 \ ,\dots, \ M_s\}$, selected as part of $\Q^*$ to receive feedback $s^j_t; \ j \in 1 \ ,\dots, \ M_s$ (model updates or gradients denoted by ${\nabla^j} \Q_{t}^{*} \in \Rc^{K \times M_s}$) from all of the users. For each item $j$, the reward $r^j_t$ is optimized by integrating the immediate and gradual rate of changes in the gradients, jointly:
\begin{equation}
	r^j_t = (\ 1 \ - \ \gamma t) \ cosine\_sim(\upsilon_t^j,{\nabla^j} \Q_{t}^*) \ + \ \frac{\gamma}{t} \sum_{k=1}^{K} \left| {\nabla^j} \Q_{t-1} - \ {\nabla^j} \Q_{t}^* \right|
	\label{eq:reward}
\end{equation}
where $\gamma$ is the regularization term. The quantities $\nabla^j \Q_{t-1}$ and  $\nabla^j \Q_{t}^*$ are the gradients of item $j$ from the $t-1$ and $t$ iterations. As stated by ADAM~\cite{kingma2015adam}, $\upsilon_t^j$ records an exponential decay of the past squared gradients for an item $j$ as: 
\begin{equation}
	\upsilon_t^j \ = \ \frac{\beta_2 \upsilon_{t-1}^j \ + \ (1 - \beta_2) \Big({\nabla^j} \Q_{t}^{*}\Big)^2}{(1 - \beta_2)}; \ \ \ \ \ \ \text{where} \ 0 < \beta_2 < 1
	\label{eq:adam_v}
\end{equation} 
Taking inspiration from stochastic gradient approaches, our method computes a composite reward regularized by the number of FL iterations. The expression $\frac{\gamma}{t} \sum_{k=1}^{K} \left|\nabla^j \Q_{t-1} - \ \nabla^j \Q_{t}^* \right|$ sums the reward as the function of the absolute differences in the gradients specifically modelling immediate changes during the initial FL iterations. The impact decreases as more rounds of updates have been completed. Whereas $(\ 1 \ - \ \gamma t) \ cosine\_sim(\upsilon_t^j,\nabla^j \Q_{t}^*)$  increases the reward as the cosine similarity of the gradual changes in the gradients increases with the increasing number of FL iterations. The composite reward supplemented by the BTS strategy aims to balance exploration and exploitation. For instance, in the beginning, the item selection depends on the rate of change in the gradients. The items whose gradients changes are large are selected more often, whereas in the later phase, the selection of items is dependent on the overall similarity of the gradients in order to favor stable convergence, particularly in the online training of the recommendation model. Moreover, the regularization parameter $\gamma$ can be tuned to adjust the strength of the information sharing between the immediate and gradual changes, scaled by the a factor $t$. For example, initializing $\gamma = 0$ restricts the method to estimate the reward by focusing on long-term gradual changes whereas $\gamma = 1$ pushes the function to infer the reward based on the immediate changes in the gradients.  
\subsection{Regret}
We believe that the regret of FCF-BTS can be bounded with respect of the FL iterations $t$. However, the existing works on FL (combined with stochastic gradient and BTS) do not provide sufficient tools for the proper analysis of our method. While the existing approaches provide $\mathcal{O}(\sqrt{NT\ln(N)})$ using Gaussian priors) regret bounds~\cite{agrawal2013further} for BTS algorithm, they do not assume the FL problem settings. Alternatively, an information-theoretic analysis proposed an entropy-based regret bound over $t$ time steps for online learning algorithm using BTS~\cite{russo2016information}.  However, their bound increases linearly according to the number of actions, which is typically large in our particular problem setting. An optimal regret bound for FCF-BTS is one that has a sub-linear dependency (or no dependency at all~\cite{dong2018information}) with the items (or \textit{--arms}), in addition to remaining sharp within the large action spaces to duly satisfy the constraints of a privacy-preserving FL recommendation environment. \\ 

\noindent To summarize, the proposed FCF-BTS method offers a number of advantages in terms of production: (i) it allows for the optimizing of the payloads without collecting the user’s private or personal information such as the user-item interactions, (ii) the optimization of the payloads is performed on the server-side, thus avoiding any additional computational overhead on the user devices, (iii) no customization is needed on the user-side, and the users perform a typical federated local model update step as part of the FRS, and (iv) it enables the smooth plug-in/out payload optimization without making changes to the FL architecture or recommendation pipeline.

\section{Related Work}\label{sec:related_work}
The payload optimization problem and our solution to it are related to communication-efficient methodologies in federated learning. We next discuss the existing methods that promote communication efficiency and relate them to our work.

\subsection{Non Recommender Systems}
For traditional FL systems, our method can be viewed as a generalized approach for effective and efficient communication at each FL round~\cite{DBLP:journals/corr/KonecnyMYRSB16} without assuming additional constraints on the users (or client devices), thus supporting privacy-sensitive applications. Several recent studies have provided practical strategies, such as the sparsification of model updates~\cite{han2020adaptive} and utilizing Golomb lossless encoding~\cite{sattler2019robust}. This is in addition to using knowledge distillation and augmentation~\cite{jeong2018federated,he2020group}, performing quantization~\cite{DBLP:journals/corr/KonecnyMYRSB16}, applying lossy compression and the dropout~\cite{caldas2018expanding}, and sub-sampling of the clients\cite{saputra2019energy}. From a theoretical perspective, these prior works have explored convergence guarantees with low-precision training in the presence of non-identically distributed data.

\textit{Federated Reinforcement Learning:} A number of recent studies have adopted reinforcement learning, primarily to address hyper-parameter optimization~\cite{NEURIPS2020_6dfe08ed} and to solve contextual linear bandits~\cite{NEURIPS2020_4311359e} in federated mode. \\

\noindent However, unlike our method, none of these methods address the key challenge of the large-scale FRS running in production, specifically the huge payloads associated with the high number of items to be recommended. 

\subsection{Recommender Systems}
Many studies have demonstrated promising results for FRS~\cite{ammad2019federated,zhou2019privacy,chai2019secure,dolui2019poster,10.1007/978-3-030-67661-2_20,qi2020privacy,tan2020federated}. The recommendation models include factorization machine and singular value decomposition~\cite{tan2020federated}, deep learning~\cite{qi2020privacy} and matrix factorization~\cite{ammad2019federated,chai2019secure,dolui2019poster}. To overcome the computation and communication costs as part of the recommendations, Chen et al.\cite{chen2018federated} extended meta-learning to federated mode. Muhammad et al.~\cite{muhammad2020fedfast} proposed a mechanism for the better sampling of users using K-means clustering and the efficient aggregation of local training models for faster convergence, hence favoring lesser communication rounds for FL model training. However, none of these approaches address the item-dependent payload optimization problem.  

Recently, Qin et al.~\cite{qin2020novel} proposed a 4-layer hierarchical framework to reduce the communication cost between the server and the users. Notably, their approach assumes that the user-item interaction behaviors (such as ratings or clicks) are public data that can be collected on a central server. The idea is to select a small candidate set of items for each user by sorting the items based on the recorded user-item interactions. Then it will transmit the user-specific candidate set to each user in order to train the local model and perform inference. Unlike theirs~\cite{qin2020novel}, our approach does not require the recording of any user sensitive interaction data and it solves the payload optimization problem in a standard federated setting with minimal computational overheads on the FL server. Our approach follows the widely accepted FRS setting without requiring any additional requirement for the users to share their sensitive data. It uses only the local model updates to solve the payload optimization problem~\footnote{Notably, we did not consider this as a baseline approach in our experiments owing to the differences in the FL architecture and the assumptions on data privacy.}. \\

\noindent To the best of our knowledge, we have proposed the first method to solve the payload optimization problem for FCF assuming an implicit feedback scenario. However, the proposed method is applicable to a wider class of FRS, particularly concerning the modelling of explicit user feedback without a loss of generality.

\section{Datasets}\label{sec:dataset}
We used three benchmark recommendation datasets to test the proposed federated payload optimization method. The datasets were processed in order to model the implicit feedback interactions in this study. The characteristics for each of the preprocessed datasets have been given in Table ~\ref{Datasets}. We dropped the \textit{--timestamp} information from the datasets,  since we only needed the user-item interactions to analyze the proposed FCF-BTS method. We selected the datasets to rigorously test FCF-BTS primarily for two reasons: (i) the datasets contain a diverse set of items ranging from 3064 to 17632, and (ii) the datasets are highly sparse in nature which is typically anticipated in a production environment. 

\subsection{Movielens-1M}
Movielens-1M \cite{harper2015movielens} rating dataset was made publicly available by the Grouplens research group \textit{(\url{https://grouplens.org/datasets/movielens/})}. The dataset contained 1,000,209 ratings of 3952 movies made by 6040 users. The rating dataset consisted of the user, movie, rating, and timestamps information. The ratings were explicit, so we converted them to implicit feedback based on the assumption that the users have watched the video that they have rated. All ratings were changed to one irrespective of their original value, and missing ratings were set to zero. 

\subsection{Last-FM}
Last-FM \cite{cantador2011second} rating dataset was made publicly available by the Grouplens research group \textit{(\url{https://grouplens.org/datasets/hetrec-2011/})}. The dataset contained 92834 listening counts of 17632 music artists by 1892 users. The listening count for each user-artist pair was set to one irrespective of the original value and missing listening counts were set to zero to convert the data into implicit feedback.

\subsection{MIND}
MIND-small \cite{wu2020mind} news recommendations dataset was made publicly available by Github \textit{(\url{https://msnews.github.io/})}. It was collected from the anonymized behavior logs of the Microsoft News website. This dataset contained the behavioral logs of 50,000 users. It was an implicit feedback dataset where 1 refers to clicked and 0 refers to non-clicked behavior. Users with at least 5 news clicks were considered. For simplicity, we denoted the MIND-small dataset with the abbreviation “MIND” throughout the manuscript.

\begin{table}[ht]
	\caption{Overview of datasets used in the study, where \#Interactions represent the total number of observed user-item interactions and Sparsity(\%) refers to the percentage of unobserved interactions in the training dataset. The Last-FM and MIND datasets are highly sparse in nature, exhibiting a similar level of sparsity to what is typically expected in production datasets.}
	\label{Datasets}      
	\centering
	\begin{tabular}{c | c c c c}
		\hline \hline
		\textbf{Datasets} &\textbf{\# Users} & \textbf{\# Items} & \textbf{\# Interactions} & \textbf{Sparsity (\%)}\\[0.5ex]
		\hline 
		Movielens-1M  & 6040 & 3064 & 914676 & 96.05\% \\
		Last-FM & 1892 & 17632 & 92834 & 99,78\% \\
		MIND & 16026 & 6923 & 163137 & 99,89\% \\
		\hline
	\end{tabular}
\end{table}

\section{Experiments} \label{sec:experiments}
To demonstrate the usefulness of the proposed bandit method, we compared the performance of FCF-BTS with three other methods. As a baseline approach, we used the FCF-Random method that does not benefit from bandits for item selection. Instead, it selects a part of the global model that is comprised of items selected at random. Furthermore, to assess the advantage of optimizing the payload in a model-driven fashion compared to the naive optimization method, we compared the FCF-BTS performance with the TopList recommendation of the most popular items to every user. 
In addition, we used FCF~\cite{ammad2019federated} as an upper-bound comparison to our FCF-BTS method. In each FL communication round, FCF (Original) transfers (uploads/downloads) the whole global model between the server and users. This provides an estimate of the recommendation performance for each dataset, achievable when no payload optimization is performed in federated mode. 
\subsection{Hyper-parameters} 
To ensure the fair treatment of all three methods, we adapted the same hyper-parameter settings for FCF (as shown in Table ~\ref{Parameters}) that were found to be optimal from the previous studies~\cite{ammad2019federated,10.1007/978-3-030-67661-2_20}. The FCF-BTS specific hyper-parameters of the prior were set as $(\mu_\theta, \tau_\theta) = (0, 10000)$ and the regularization of the reward was set as $\gamma = 0.999$. 
\begin{table}[ht]
	\caption{FCF's hyper-parameters values used in our experiments. K represents the number of latent factors, $\lambda$ is the L2-regularization term, $\alpha$ denotes the implicit confidence parameter. $\beta_1$, $\beta_2$, $\eta$, and $\epsilon$ are the parameters of the ADAM optimizer.}
	\label{Parameters}
	\centering
	\begin{tabular}{c c c c c c c c}
		\hline \hline
		Model & K & $\lambda$ & $\alpha$ & $\beta_1$ & $\beta_2$ & $\eta$  & $\epsilon$ \\[0.5ex]
		\hline
		FCF  & 25 & 1 & 4 & 0.1 & 0.99 & 0.01 & 1e-8 \\
		\hline
	\end{tabular}
\end{table}
\newline
The threshold parameter $\Theta$ in Algorithm~\ref{alg:fl_payload} refers to the number of federated model updates that are needed to update the global model.   
For each dataset, we selected $\Theta$ as (Movielens, Last-FM, MIND) $ = (100, 100, 500)$ relative to the total number of users~\cite{ammad2019federated,10.1007/978-3-030-67661-2_20}.

\subsection{Model training and evaluation criteria}
We followed the training and evaluation approach of Flanagan et al.~\cite{10.1007/978-3-030-67661-2_20} and performed 3 rounds of model rebuilds. The training set of every user was comprised of 80\% item interactions that were selected at random. The performance metrics were then computed on the remaining 20\% of interactions (test set) for each user separately. Likewise, the users’ performance metrics were also aggregated to update the global metric values on the FL server.
Notably, the FL server triggers the update of the global model if the \textit{$\mbox{NumberGradientUpdates}>=\mbox{$\Theta$}$}, implying that in each iteration, only a subset of users sent their test set performance metrics along with the local model updates. At the 1000th iteration, we took the average of the previous ten global metric values to account for the biases that originate from the unequal test set distributions of the users sending asynchronous updates to the FL server. 

We used well-known recommendation metrics \cite{bobadilla2013recommender} namely Precision, Recall, F1, and Mean Average Precision (MAP) to evaluate our models for top $10$ predicted recommendations, given the recommendation list length of 100. To implement these metrics, we adapted the formulation of Flanagan et al.~\cite{10.1007/978-3-030-67661-2_20} (as described in their equations S2 - S5). To make the recommendation metrics comparable, we further normalized the performance metrics using the theoretically best achievable metrics for each dataset. We computed the theoretically best metrics by recommending items from the test set of each user. However, if the user had less than 100 items in their test set, a recommendation list was formed by adding items at random with which the user has not interacted with in the past. Likewise, the TopList performance metrics were estimated using the 100 most popular items ranked by their interaction frequency in the training set. 

Finally, we calculated two summary statistics to analyze the effect of the payload reduction on the recommendation performance degradation namely ``$\textbf{Impr \%}$" to quantify the relative performance improvement of FCF-BTS compared to FCF-Random or TopList, and ``$\textbf{Diff \%}$" to compute the relative difference between FCF-BTS and FCF (Original) performances,
\begin{equation}
	Impr \% = \Big| \frac{Metric_{Mean}^{FCF-BTS}\ - \ Metric_{Mean}^{FCF-Random/TopList}}{Metric_{Mean}^{FCF-Random/TopList}} \Big| * 100
\end{equation}

\begin{equation}
	Diff \% = \Big | \frac{Metric_{Mean}^{FCF-BTS}\ - \ Metric_{Mean}^{FCF}}{Metric_{Mean}^{FCF}} \Big| * 100
\end{equation}

\noindent where $Metric_{Mean}^{(.)}$ is the mean of the recommendation metric values across 3 model builds. 

\section{Results}\label{sec:results}
As FCF-BTS is the first payload optimization method for FRS, we used FCF-Random and TopList as the baseline comparison methods. We rigorously analyzed the effect of payload reduction on the recommendation performance degradation (loss of accuracy) using FCF-BTS and FCF-Random. In particular, we analyzed the recommendation performance when 25\%, 50\%, 75\%, 80\%, 85\%, 90\%, 95\% or 98\% of the original model payload was reduced. In practice, this payload reduction implies that 75\%, 50\%, 25\%, 20\%, 15\%, 10\%, 5\% or 2\%-of-items from the total number of items has been used during the FL model training. 

\begin{figure}
	\centering
	\begin{center}
		Movielens
	\end{center}
	\includegraphics[width=0.24\textwidth]{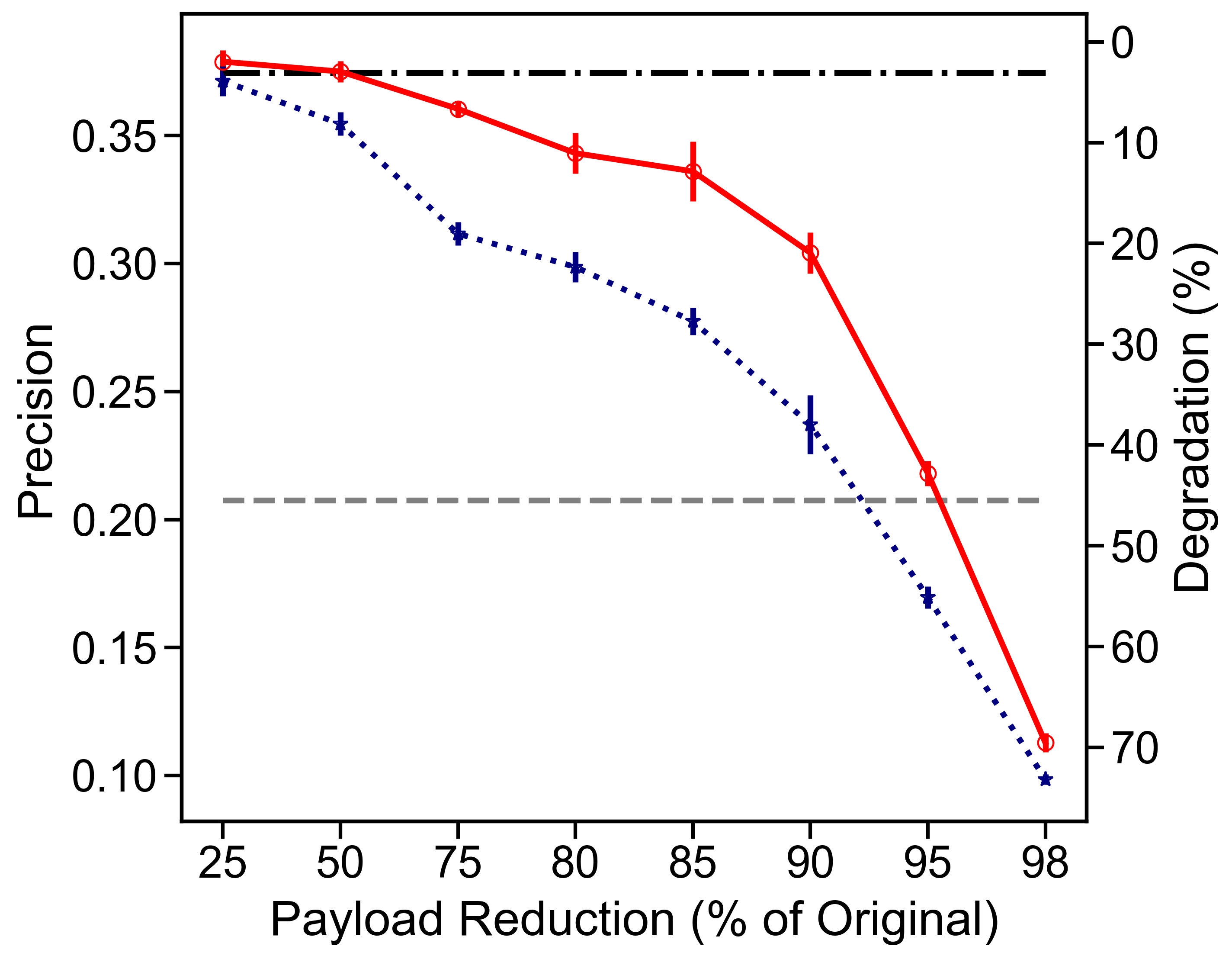}
	\includegraphics[width=0.24\textwidth]{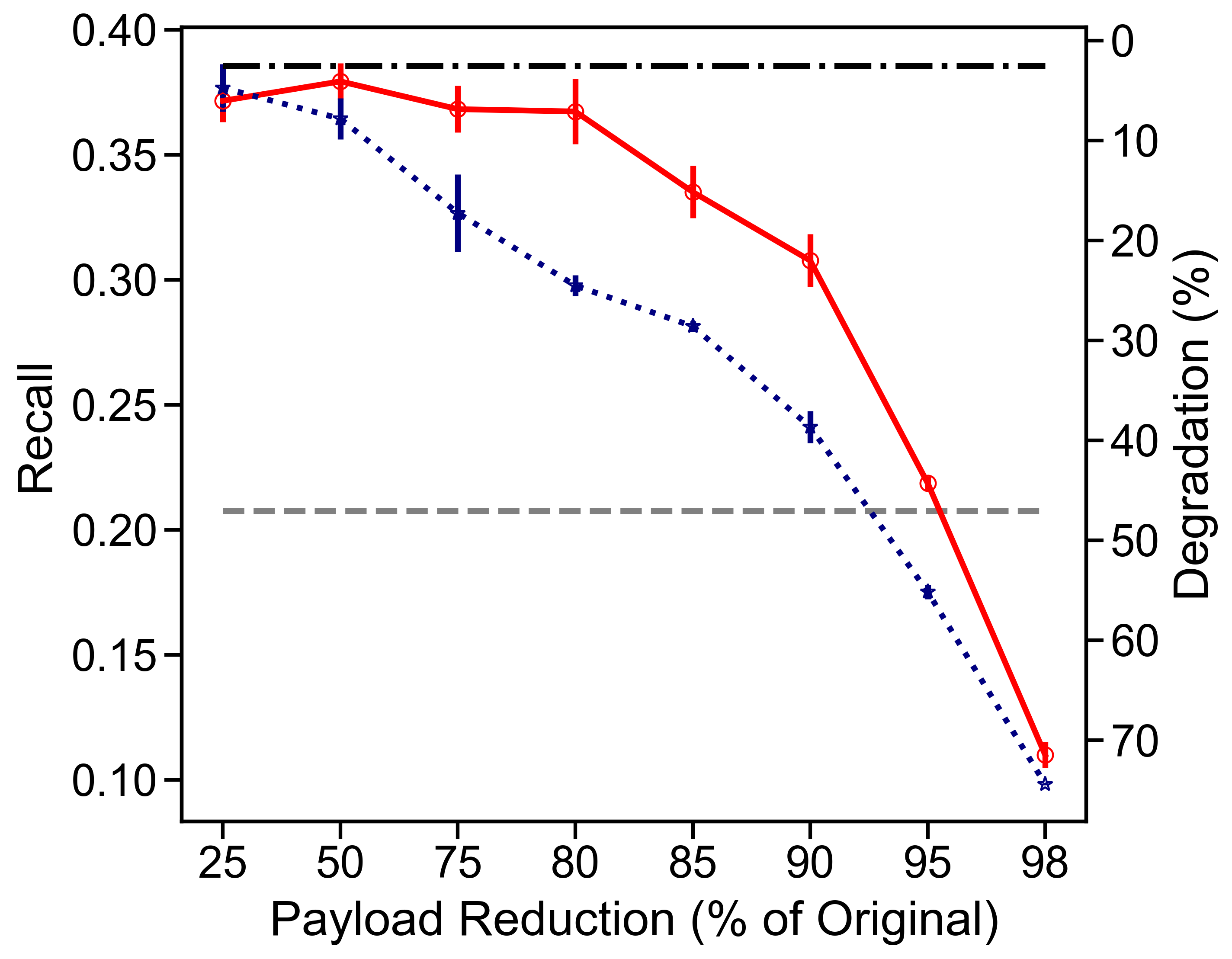}
	\includegraphics[width=0.24\textwidth]{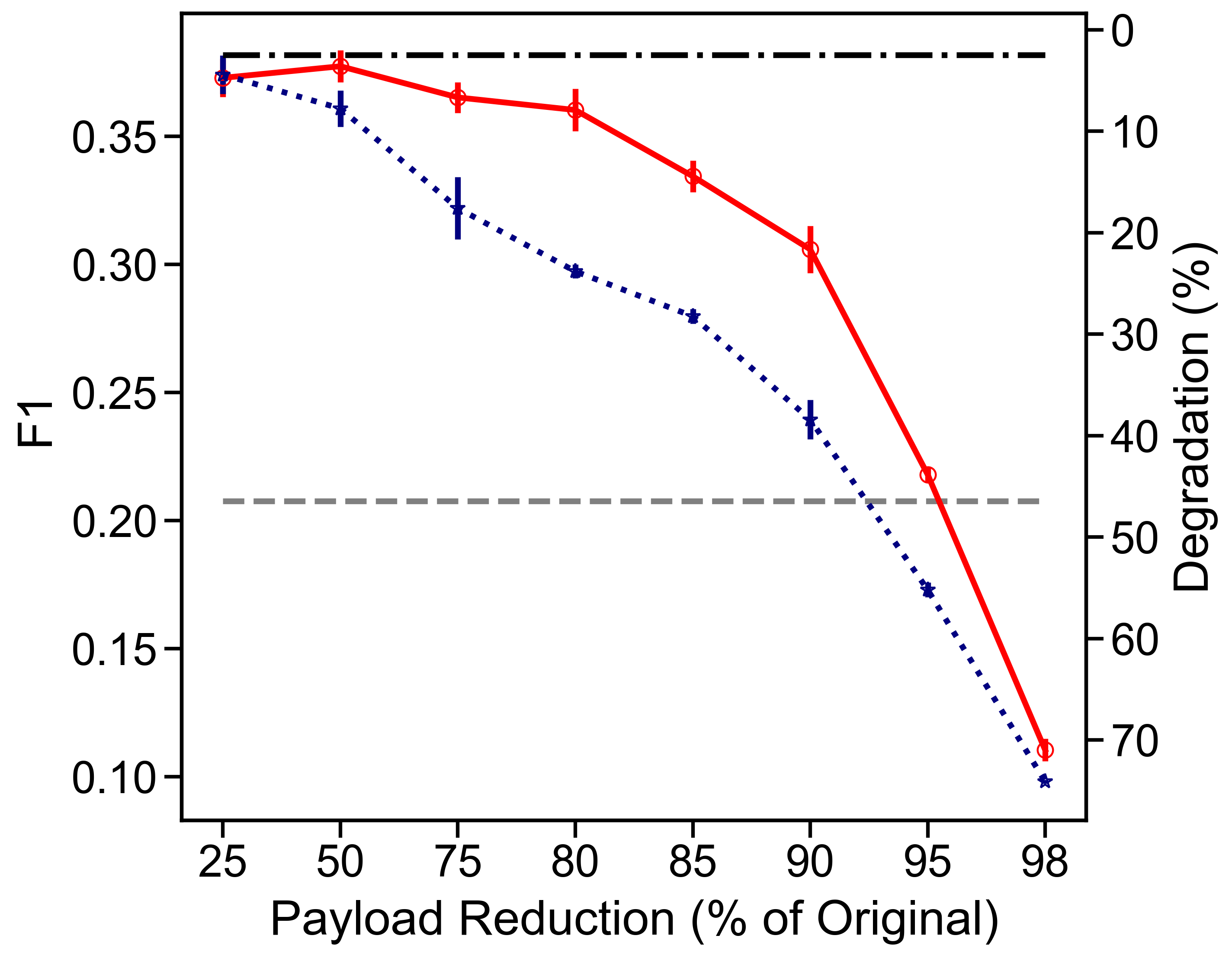}
	\includegraphics[width=0.24\textwidth]{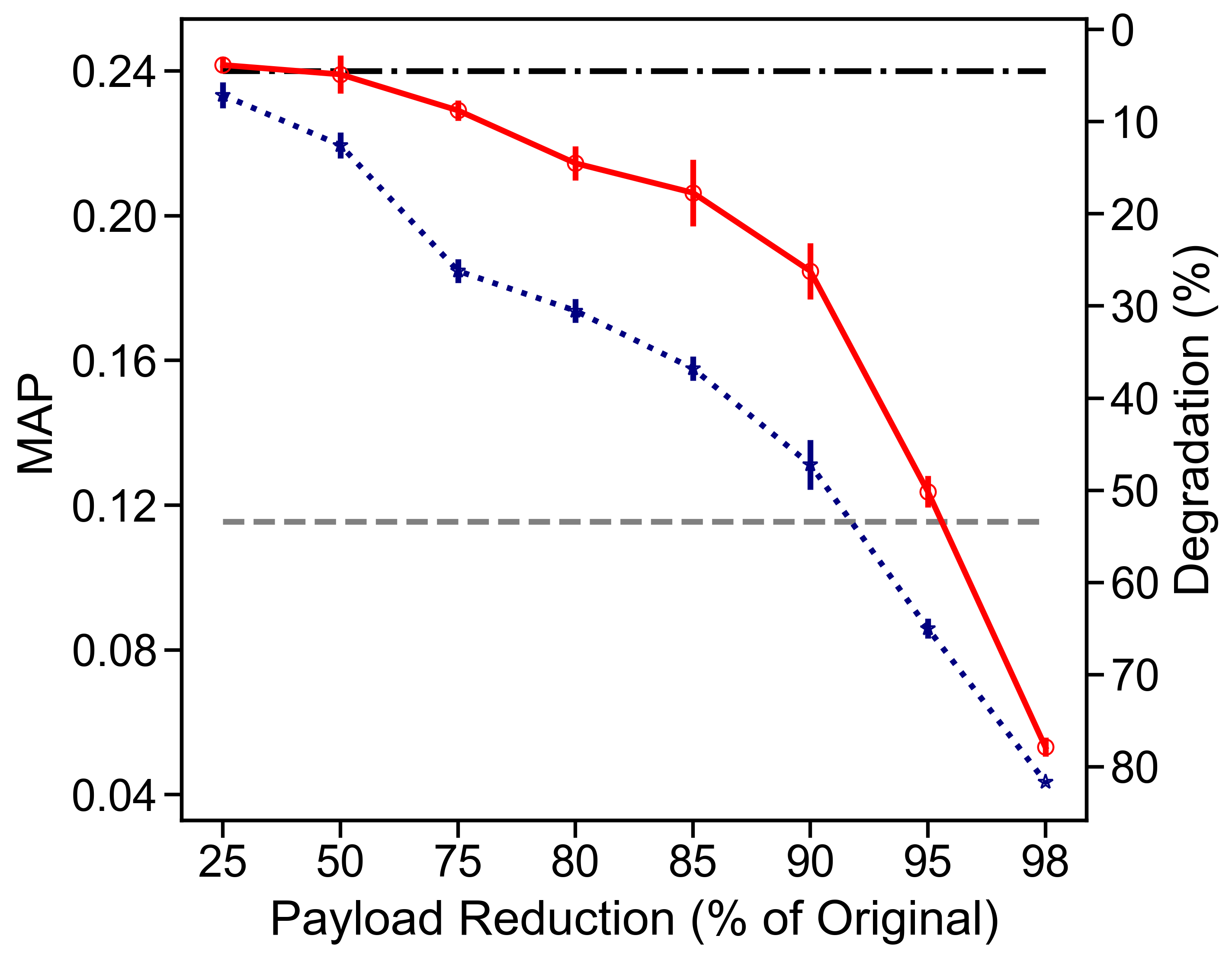}
	\begin{center}
		Last-FM
	\end{center}
	\includegraphics[width=0.24\textwidth]{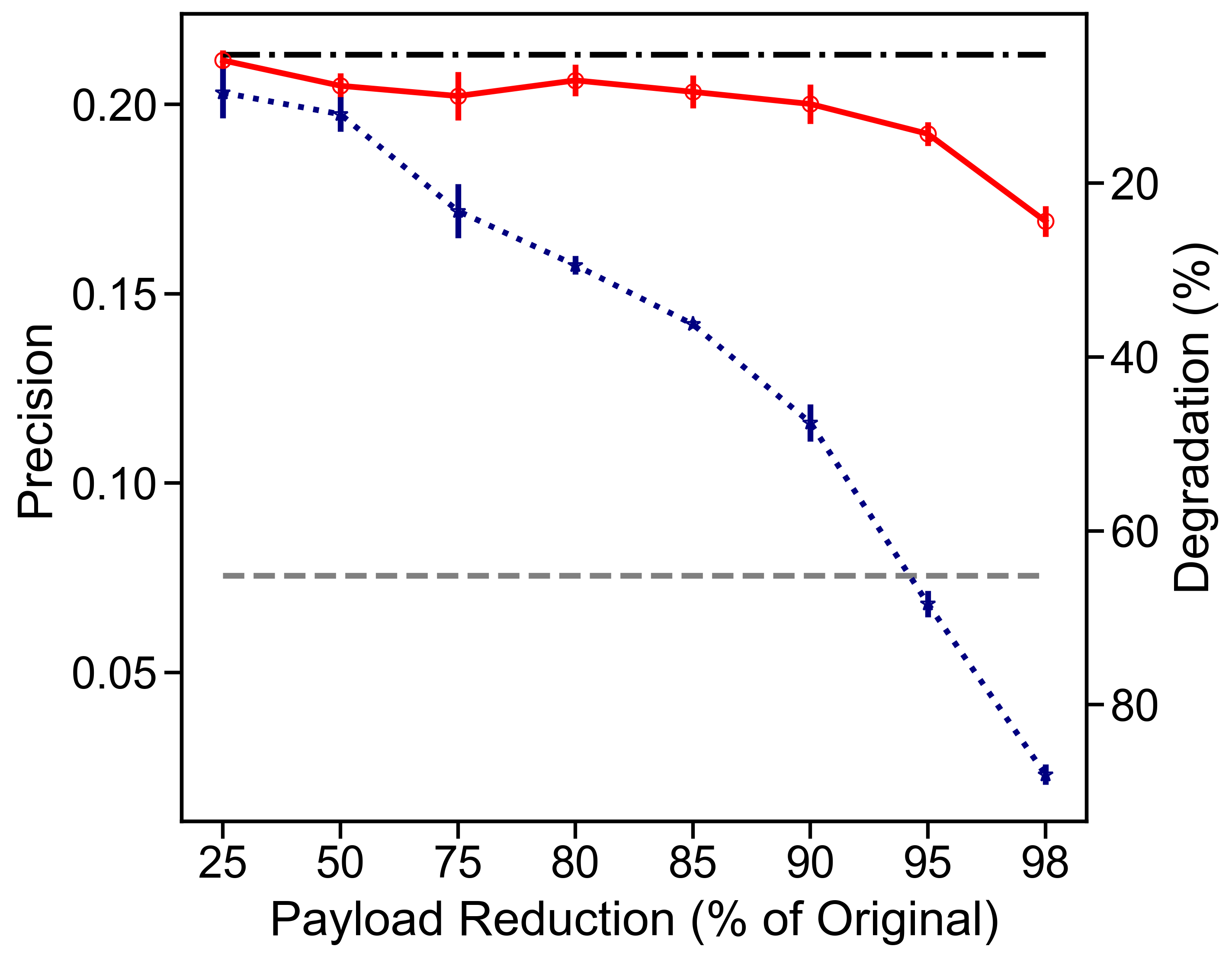}
	\includegraphics[width=0.24\textwidth]{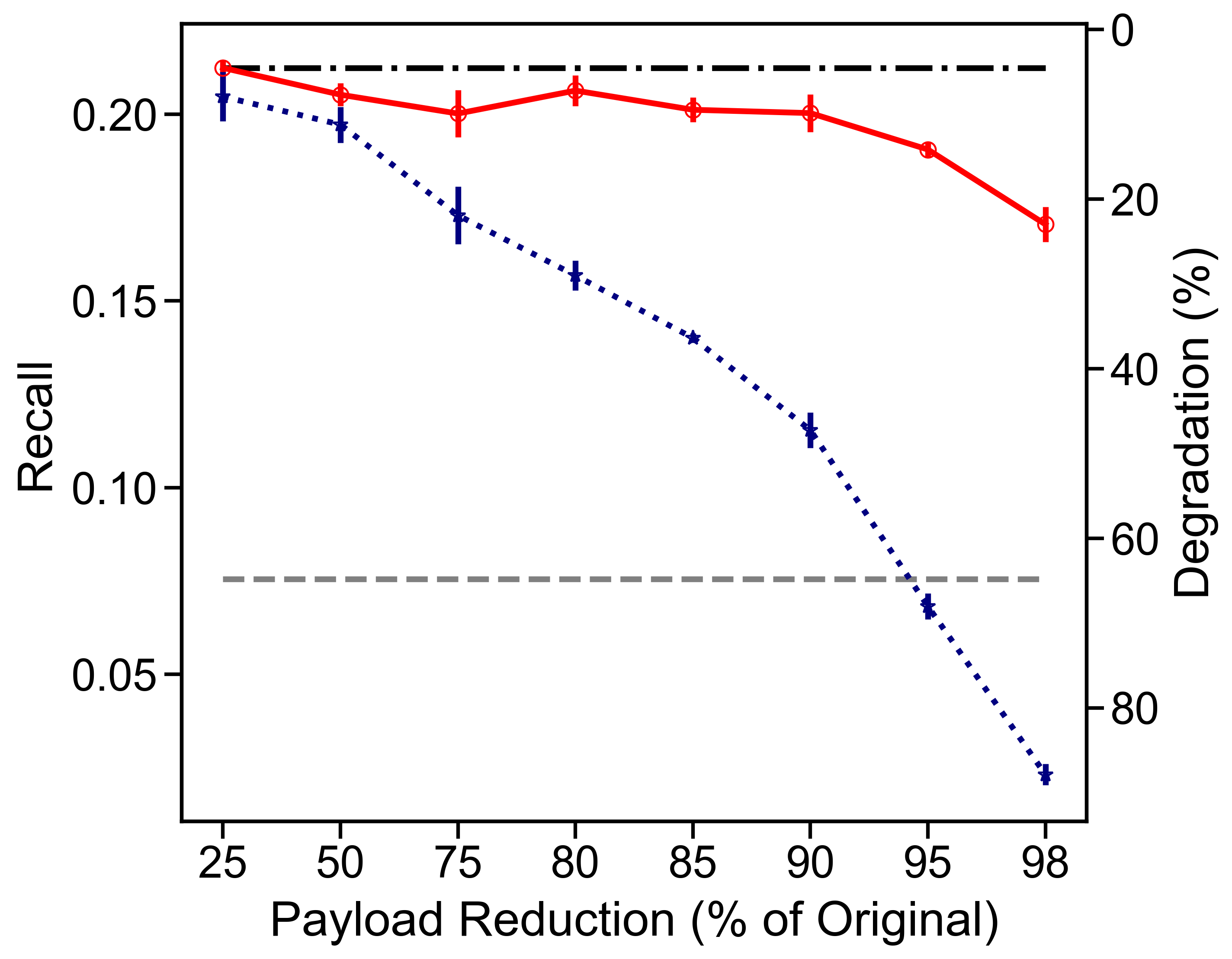}
	\includegraphics[width=0.24\textwidth]{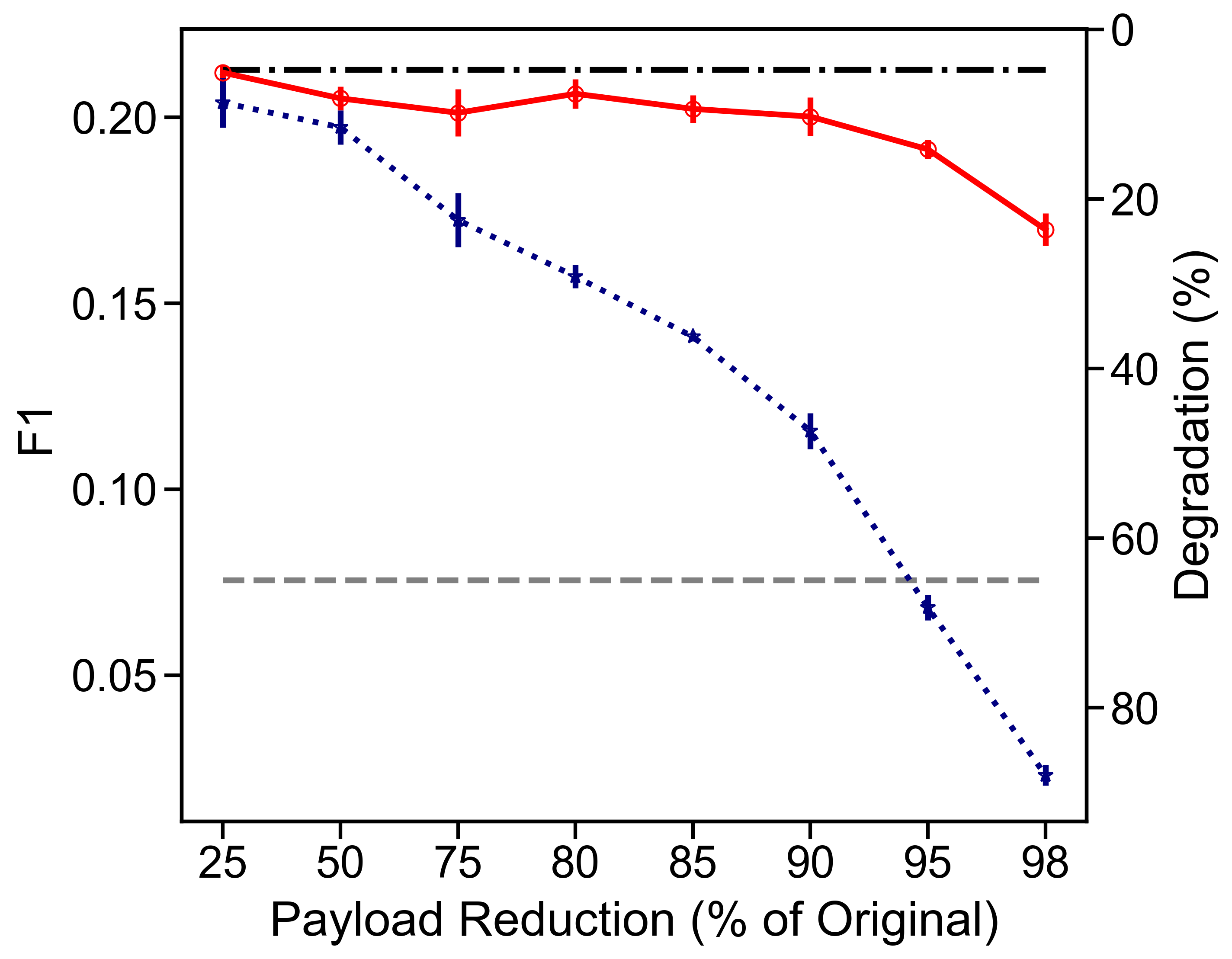}
	\includegraphics[width=0.24\textwidth]{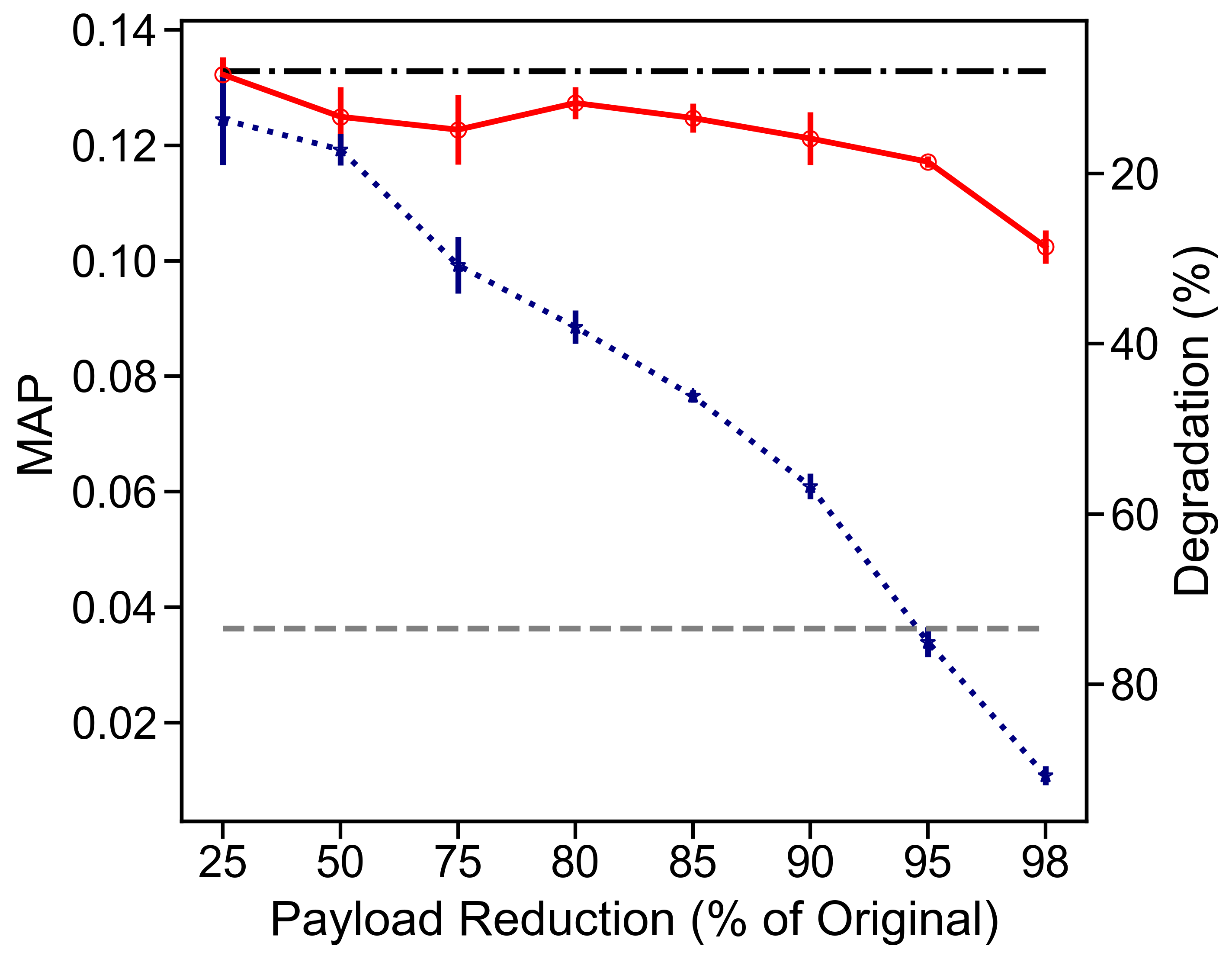}
	\begin{center}
		MIND
	\end{center}
	\includegraphics[width=0.24\textwidth]{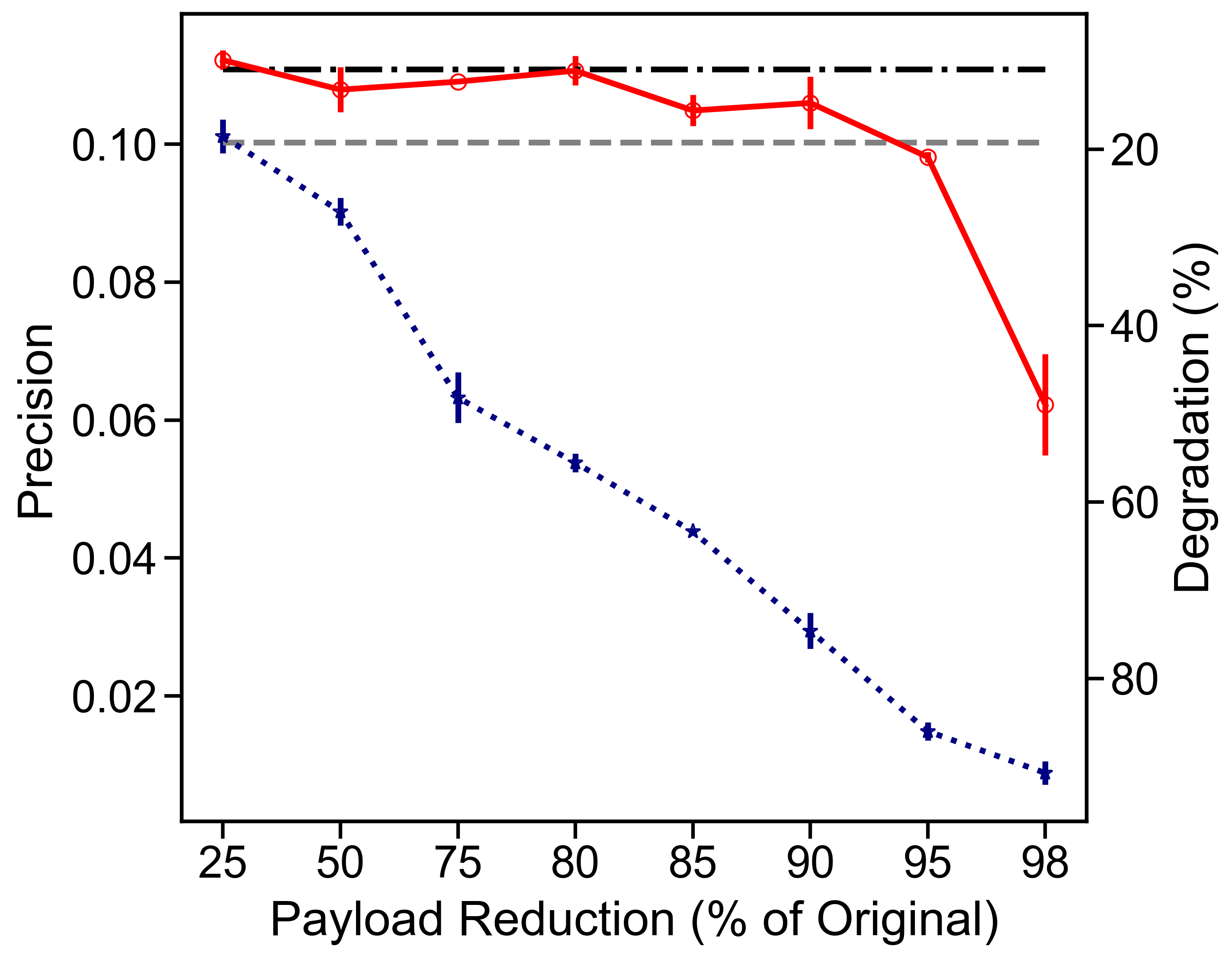}
	\includegraphics[width=0.24\textwidth]{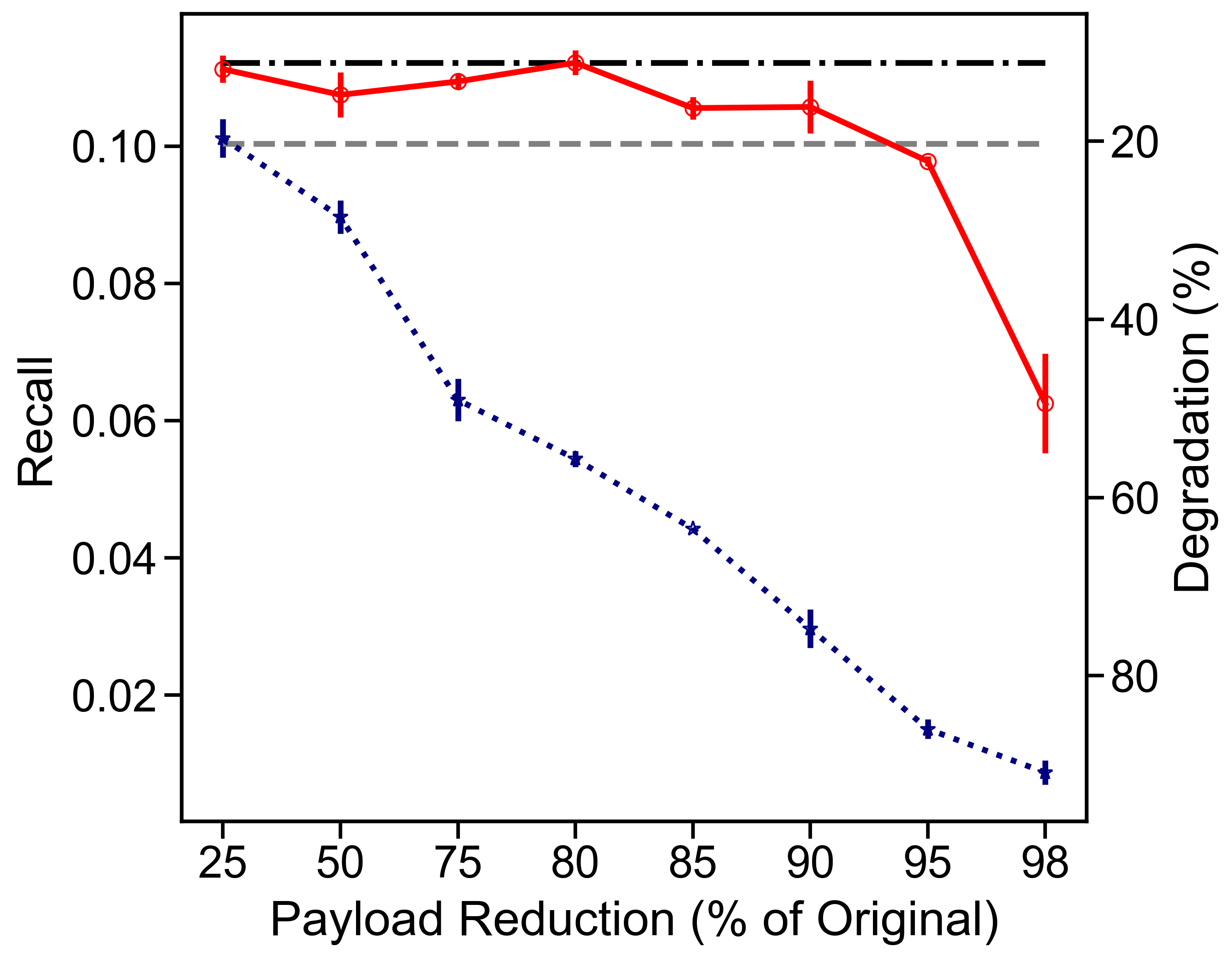}
	\includegraphics[width=0.24\textwidth]{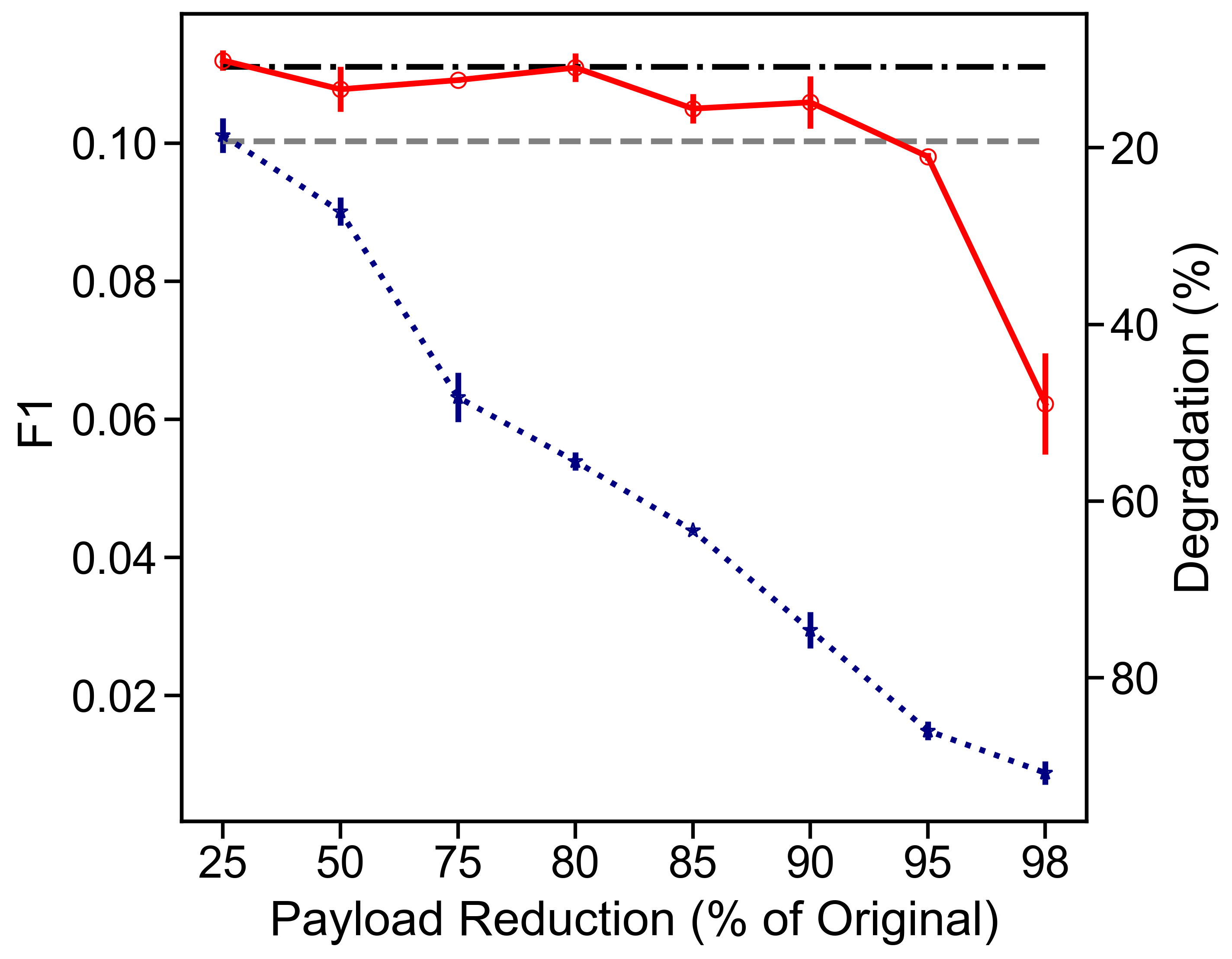}
	\includegraphics[width=0.24\textwidth]{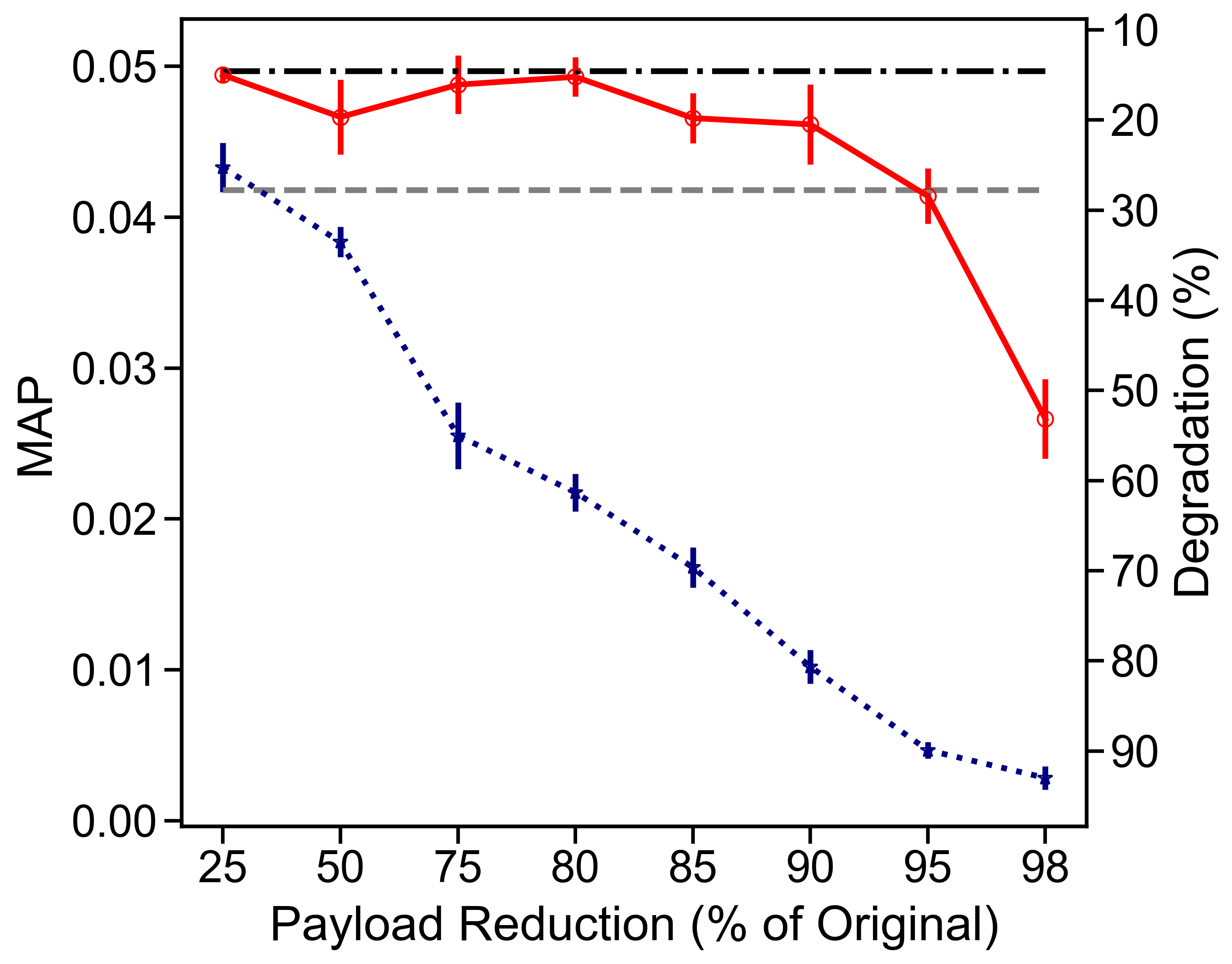}
	\includegraphics[width=0.5\textwidth]{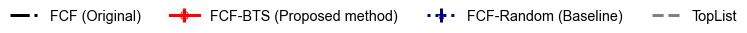}
	
	\caption{The effects of payload reduction on the recommendation performance degradation (loss of accuracy). The X-axis denotes the \% reduction in payload of the original model. The Y-axis (left side) represents the metric values, while the \% degradation compared to the original model’s performance is shown on the right side. Each point denotes the average test set recommendation performance over three rounds of model rebuild with error bars showing the standard deviation over the mean. The proposed FCF-BTS consistently outperforms FCF-Random (baseline) and demonstrates a substantial performance gain compared to the TopList recommendations while minimizing the payload by up to 90\%. }
	\label{Performance}
	
\end{figure}

The results demonstrate that the FCF-BTS outperforms FCF-Random (Baseline) consistently as shown in Figure~\ref{Performance}. We noticed a significant improvement for highly sparse datasets such as Last-FM and MIND. In comparison to the upper-bound method, FCF-BTS closely matches the performance of FCF (Original) in the Last-FM and MIND datasets as there was up to a 90\% reduction in the model payload, confirming that FCF-BTS achieves the required performance with an extremely small payload. The method gets close in the Movielens dataset with a 75\% payload reduction. This finding implies that the use of bandits is beneficial for production datasets that are inherently sparse in nature. Most importantly, FCF-BTS yields substantial performance gains compared to the TopList recommendations in the Last-FM dataset while using only 2\% of the model payload. It shows a comparable performance in Movielens and MIND when 5\% of the items are used for model training. 

Particularly, FCF-BTS showed promising results with a 90\% payload reduction for all three datasets as shown in Table ~\ref{ResultsTable}. In the Movielens dataset, the performance degradation for precision, recall, F1 and MAP was  18.77\%, 20.19\%, 19.88\% and 23.06\% respectively, compared to the performance achievable by the FCF (original) model. On the other hand, FCF-BTS improved precision (28.3\%), recall (27.57\%), F1 (27.74\%) and MAP (40.75\%) relative to FCF-Random (Baseline) and similarly, FCF-BTS showed precision (46.53\%), recall (48.19\%), F1 (47.32\%) and MAP (59.99\%) incremental improvements compared to the TopList recommendations. 

In the Last-FM dataset, FCF-BTS had 6.12\%, 5.69\%, 5.93\% and 8.8\% less precision, recall, F1 and MAP metrics respectively, compared to the upper-bound performance metrics. FCF-BTS showed an increase in precision (72.64\%), recall (73.6\%), F1 (73.1\%) and MAP (98.85\% ) over FCF-Random (Baseline).  In comparison to the TopList, FCF-BTS resulted in substantially better recommendations while improving precision, recall, F1, and MAP by  164.88\%, 165.14\% 164.93\% and 233.44\% respectively (see Table ~\ref{ResultsTable}).
\begin{table}[htbp]
	\centering
	\small
	\caption{A detailed analysis of the 90\% payload reduction for the recommendation performance degradation (loss of accuracy). The values denote the mean ± standard deviation of the test set recommendation performance across 3 rounds of model rebuild. “Diff (\%)" represents the relative percentage differences in the performance of the proposed FCF-BTS model compared to the upper-bound performance achievable by FCF (Original). “Impr (\%)" refers to the relative percentage improvements in the performance of the proposed FCF-BTS model in comparison to the FCF-Random (Baseline) and TopList recommendations. Notably, FCF-BTS consistently outperforms both of the baseline methods including FCF-Random (Baseline) and TopList. With a $\sim$4\%--8\% loss in accuracy, FCF-BTS closely matches the performance of FCF (Original) when it comes to highly sparse datasets (Last-FM and MIND). }
	\label{ResultsTable}%
	\resizebox{0.9\columnwidth}{!}{%
		\begin{tabular}{@{}lllll@{}}
			\hline 
			& Precision & Recall & F1    & MAP  \\ [0.5ex]
			\hline
			\multicolumn{5}{c}{\textbf{Movielens-1M}}  \\
			\hline 
			FCF & 0.3744$\pm$0.00582 & 0.3855$\pm$0.00754 & 0.3817$\pm$0.00566 &  0.2400$\pm$0.00702 \\
			\textbf{FCF-BTS} & \textbf{0.3041$\pm$0.00801} & \textbf{0.3076$\pm$0.01055} & \textbf{0.3058$\pm$0.00918} & \textbf{0.1846$\pm$0.00774} \\
			FCF-Random & 0.2370$\pm$0.01154 & 0.2411$\pm$0.00644 & 0.2394$\pm$0.00765 & 0.1311$\pm$0.00685 \\
			TopList & 0.2075$\pm$0.00027 & 0.2076$\pm$0.00052 & 0.2076$\pm$0.00046 & 0.1154$\pm$0.00014 \\
			\hline
			\textbf{FCF-BTS} vs. FCF (Diff\%) & 18.77 & 20.19 & 19.88 & 23.06 \\
			\textbf{FCF-BTS} vs. FCF-Random (Impr\%) & 28.3  & 27.57 & 27.74 & 40.75\\
			\textbf{FCF-BTS} vs. TopList (Impr\%) & 46.53 & 48.19 & 47.32 & 59.99 \\
			\hline
			\multicolumn{5}{c}{\textbf{Last-FM}}  \\
			\hline
			FCF  & 0.2131$\pm$0.01128 & 0.2124$\pm$0.01044 & 0.2127$\pm$0.01086 & 0.1328$\pm$0.00745 \\
			\textbf{FCF-BTS}   & \textbf{0.2001$\pm$0.00523} & \textbf{0.2003$\pm$0.00502} & \textbf{0.2001$\pm$0.00512} & \textbf{0.1211$\pm$0.00456} \\
			FCF-Random & 0.1159$\pm$0.00487 & 0.1153$\pm$0.00479 & 0.1156$\pm$0.00482 & 0.0609$\pm$0.00218 \\
			TopList & 0.0755$\pm$0.00233 & 0.0755$\pm$0.00232 & 0.0755$\pm$0.00232 & 0.0363$\pm$0.00139 \\		
			\hline
			\textbf{FCF-BTS} vs. FCF (Diff\%) & 6.12  & 5.69  & 5.93  & 8.8  \\
			\textbf{FCF-BTS} vs. FCF-Random (Impr\%) & 72.64 & 73.6  & 73.1  & 98.85\\
			\textbf{FCF-BTS} vs. TopList (Impr\%) & 164.88 & 165.14 & 164.93 & 233.44 \\
			\hline
			\multicolumn{5}{c}{\textbf{MIND}}  \\
			\hline
			FCF  & 0.1108$\pm$0.00314 & 0.1121$\pm$0.00438 & 0.1110$\pm$0.00339 & 0.0496$\pm$0.00286\\
			\textbf{FCF-BTS}   & \textbf{0.1059$\pm$0.00379} & \textbf{0.1057$\pm$0.00386} & \textbf{0.1059$\pm$0.00380} & \textbf{0.0461$\pm$0.00264} \\
			FCF-Random & 0.0294$\pm$0.00259 & 0.0296$\pm$0.00281 & 0.0294$\pm$0.00263 & 0.0102$\pm$0.00112 \\
			TopList & 0.1002$\pm$0.00067 & 0.1003$\pm$0.00046 & 0.1003$\pm$0.00063 & 0.0418$\pm$0.00044 \\
			\hline
			\textbf{FCF-BTS} vs. FCF (Diff\%) & 4.43  & 5.71  & 4.67  & 7.1  \\
			\textbf{FCF-BTS} vs. FCF-Random (Impr\%) & 260.06 & 256.1 & 259.32 & 352.46 \\
			\textbf{FCF-BTS} vs. TopList (Impr\%) & 5.67  & 5.32  & 5.58  & 10.39 \\
			\hline
		\end{tabular}%
	}	
\end{table}%
Lastly, for the MIND dataset, the performance of FCF-BTS closely matched the performance achievable by the FCF (Original) model. The relative differences in precision, recall, F1 and MAP metrics were 4.43\%, 5.71\%, 4.67\% and 7.1\% respectively, which are small compared to the performance differences given by FCF-Random (Baseline). FCF-BTS significantly outperformed FCF-Random (Baseline) with 260.06\%, 256.1\%, 259.32\% and 352.46\% higher precision, recall, F1 and MAP metrics, respectively.  In contrast to the performance of TopList, FCF-BTS demonstrates incremental increases in precision (5.67\%), recall (5.32\%), F1 (5.58\%) and MAP (10.39\%). 

\begin{figure}
	\centering
	\begin{center}
		Movielens
	\end{center}
	\includegraphics[width=0.24\textwidth]{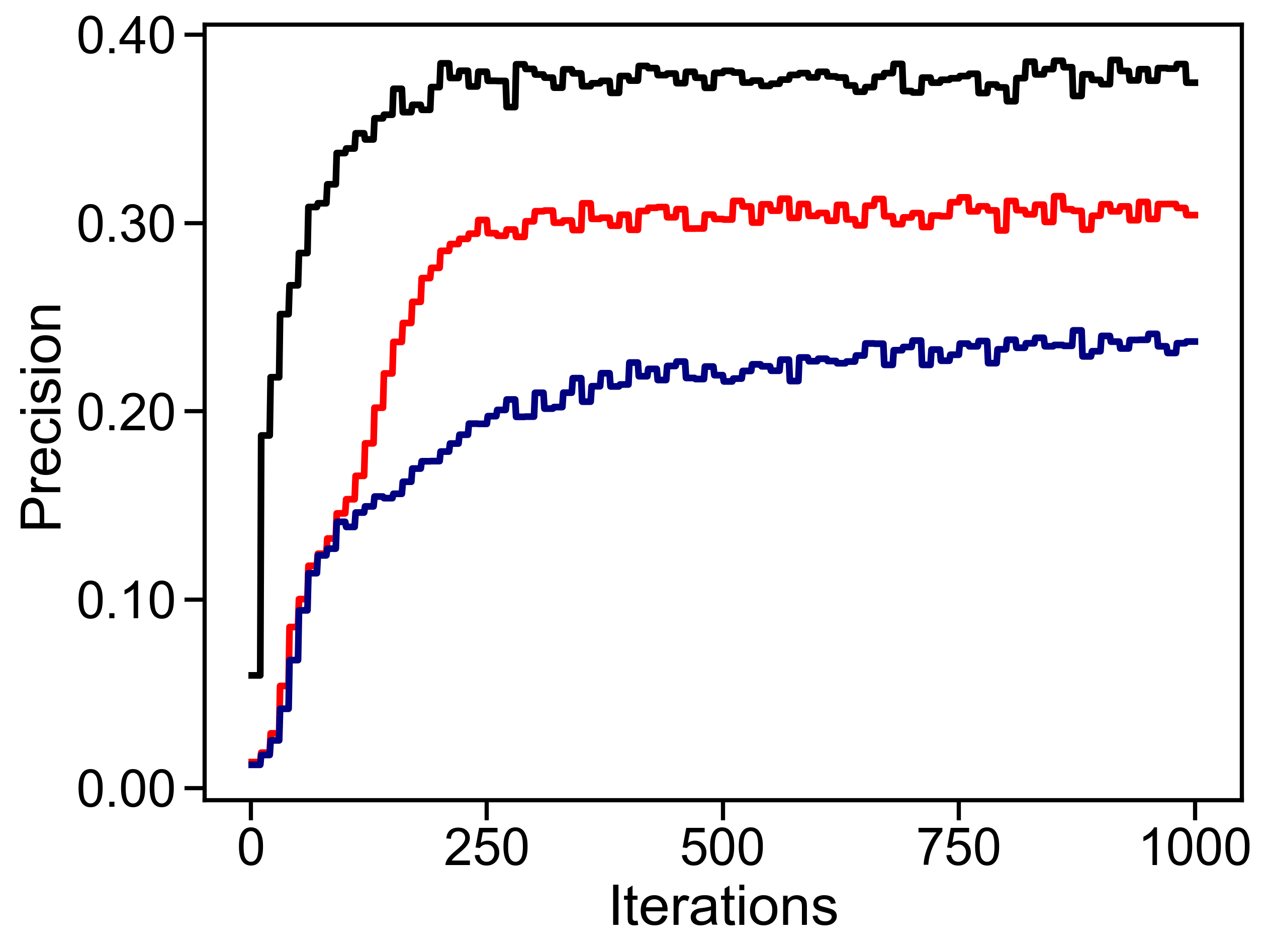}
	\includegraphics[width=0.24\textwidth]{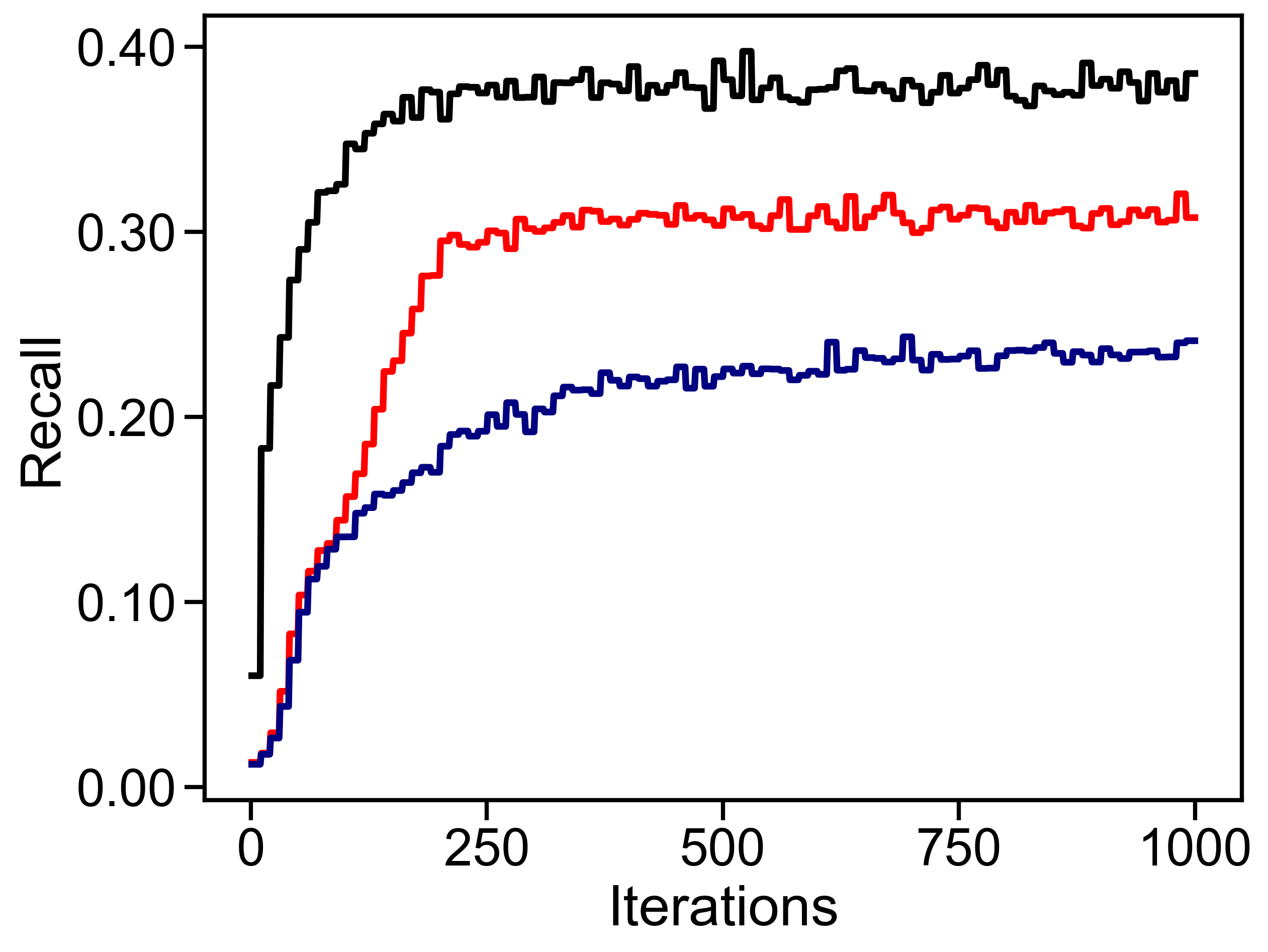}
	\includegraphics[width=0.24\textwidth]{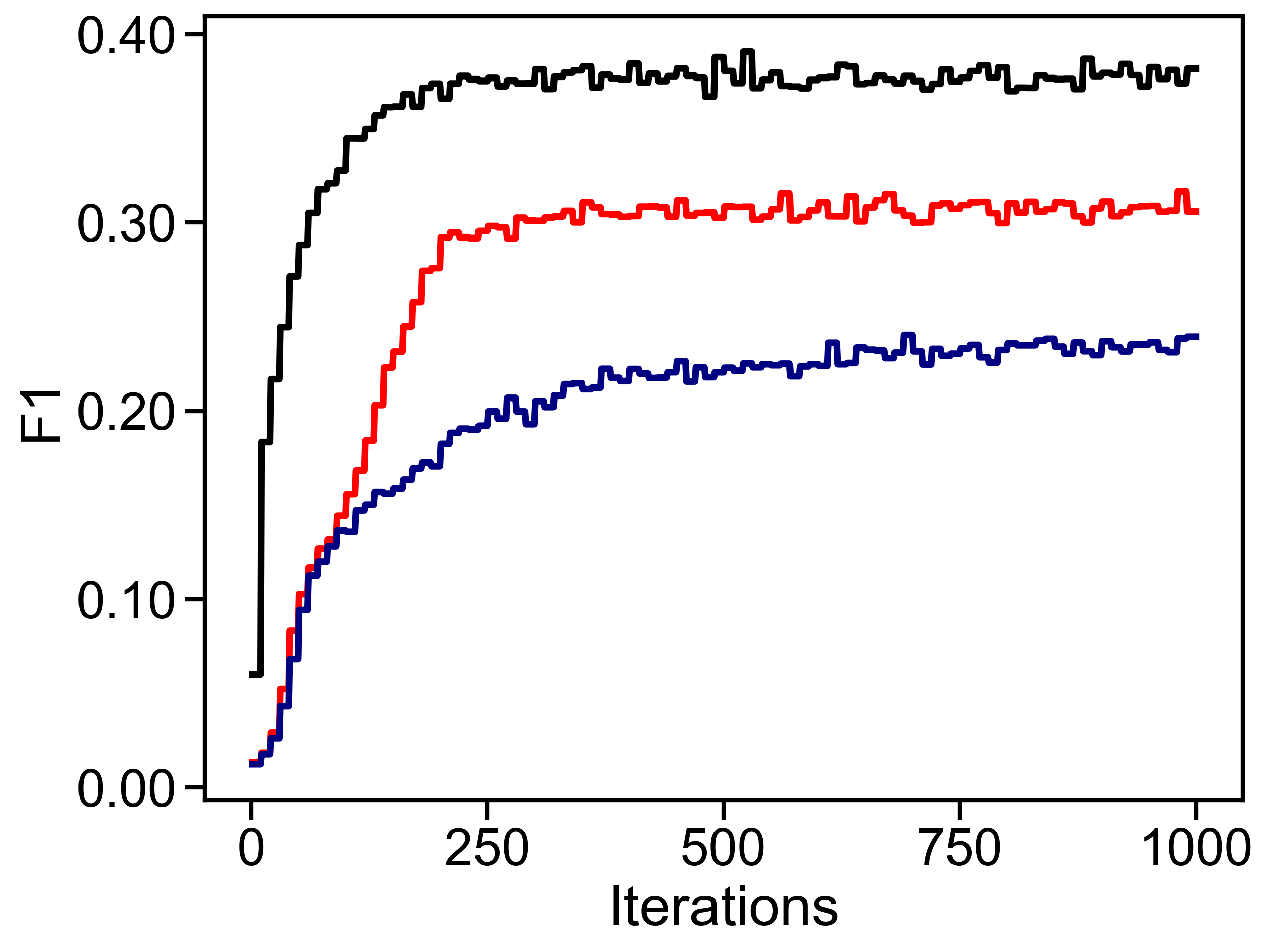}
	\includegraphics[width=0.24\textwidth]{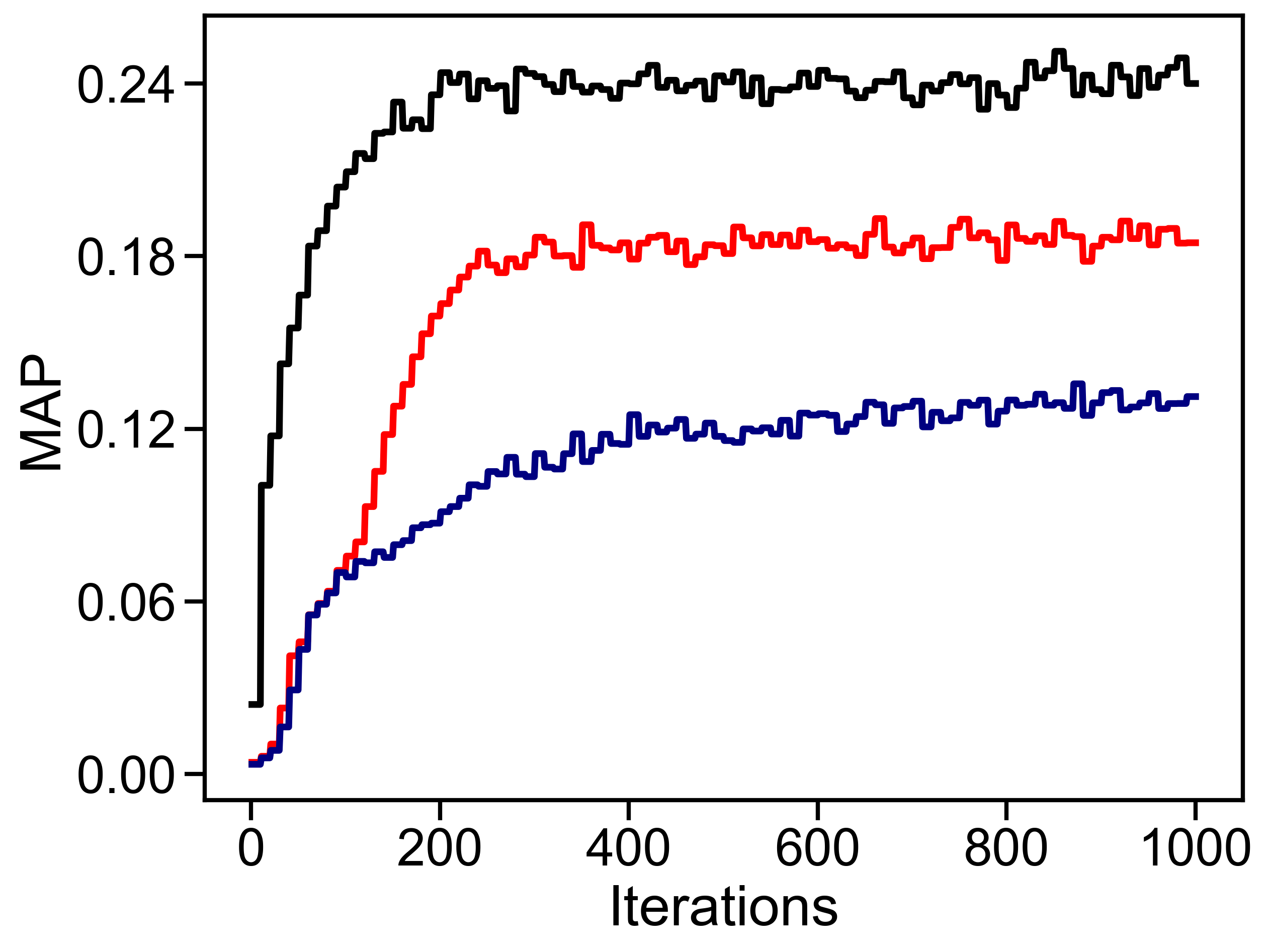}
	\begin{center}
		Last-FM
	\end{center}
	\includegraphics[width=0.24\textwidth]{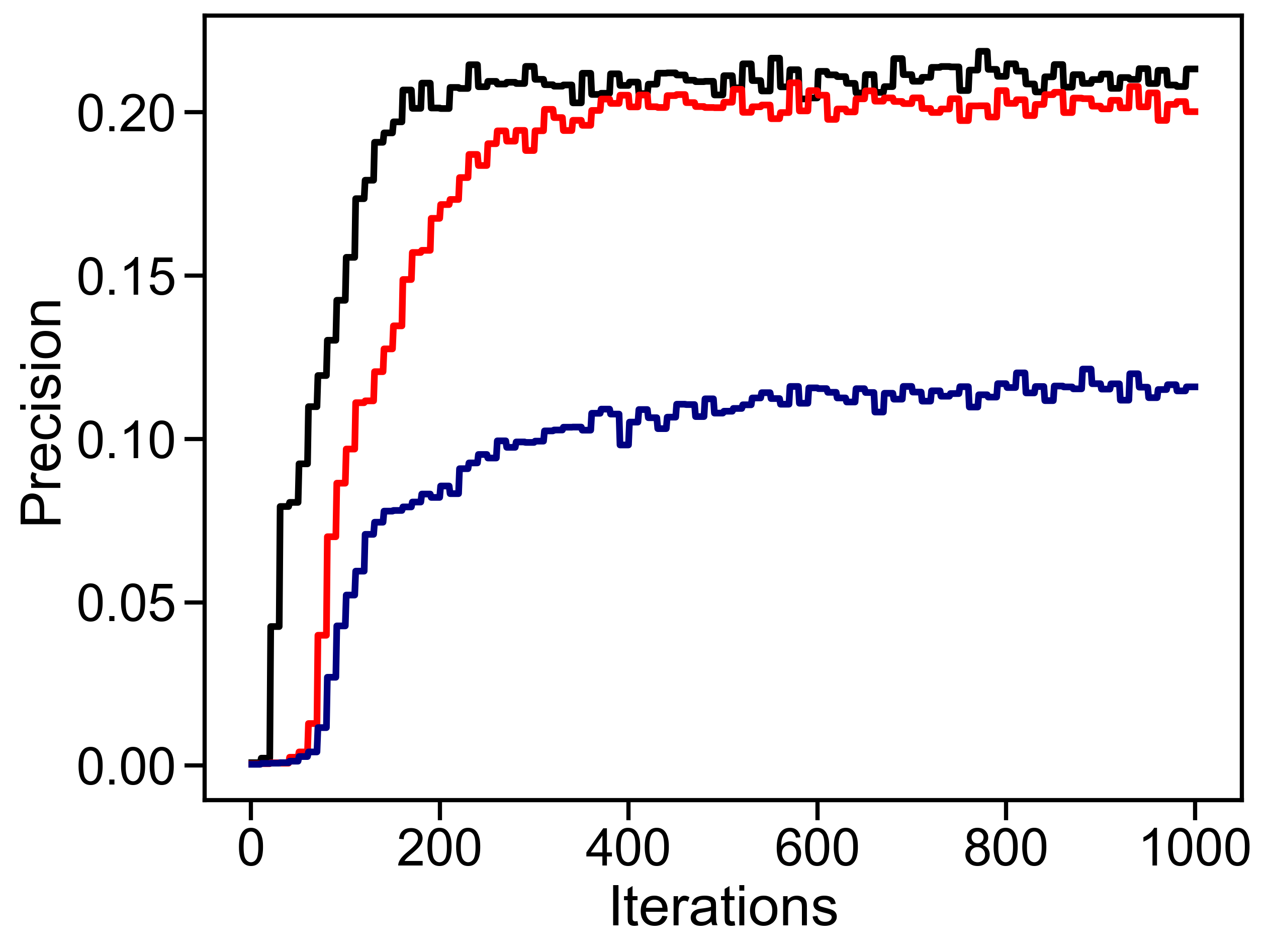}
	\includegraphics[width=0.24\textwidth]{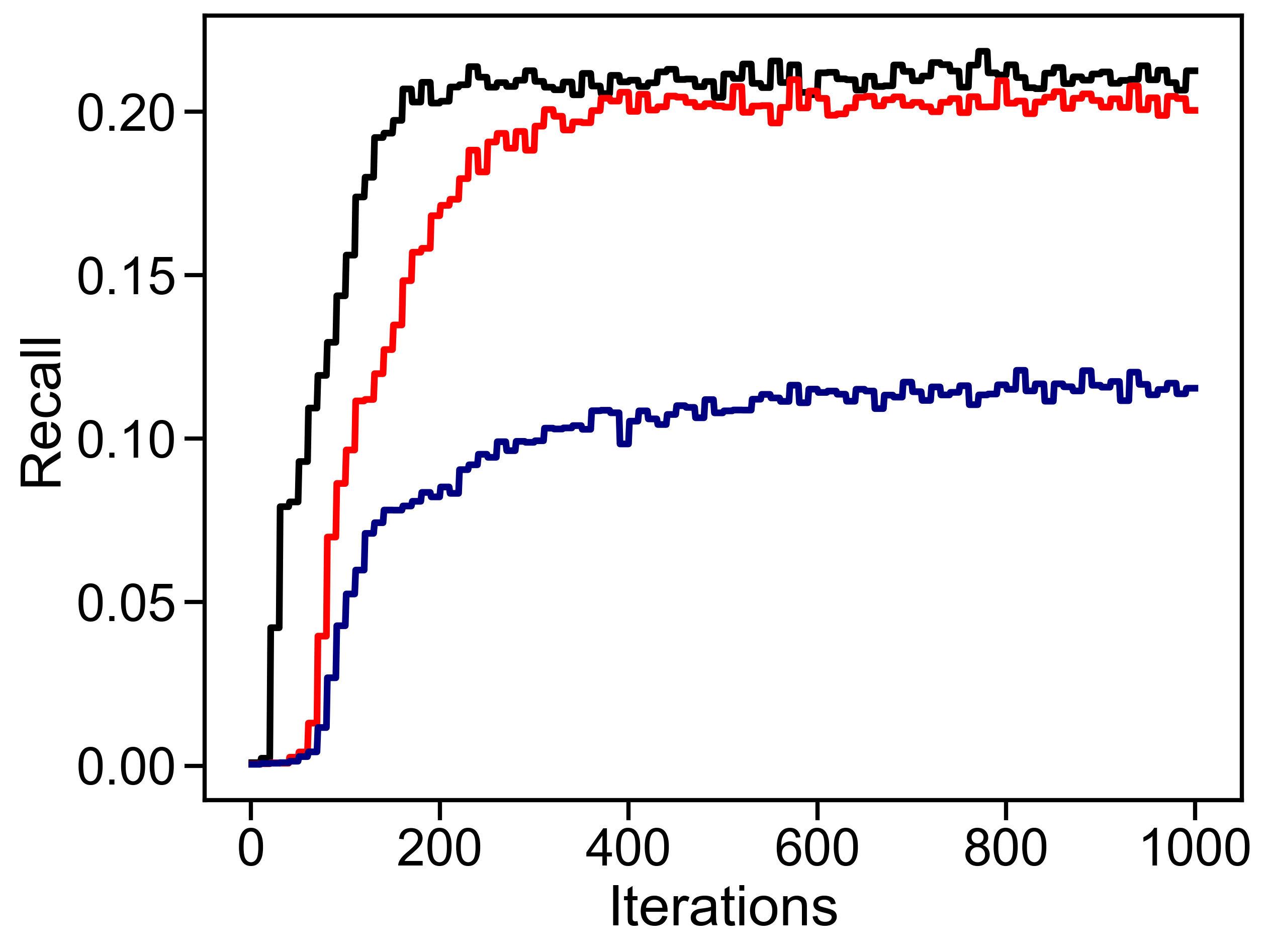}
	\includegraphics[width=0.24\textwidth]{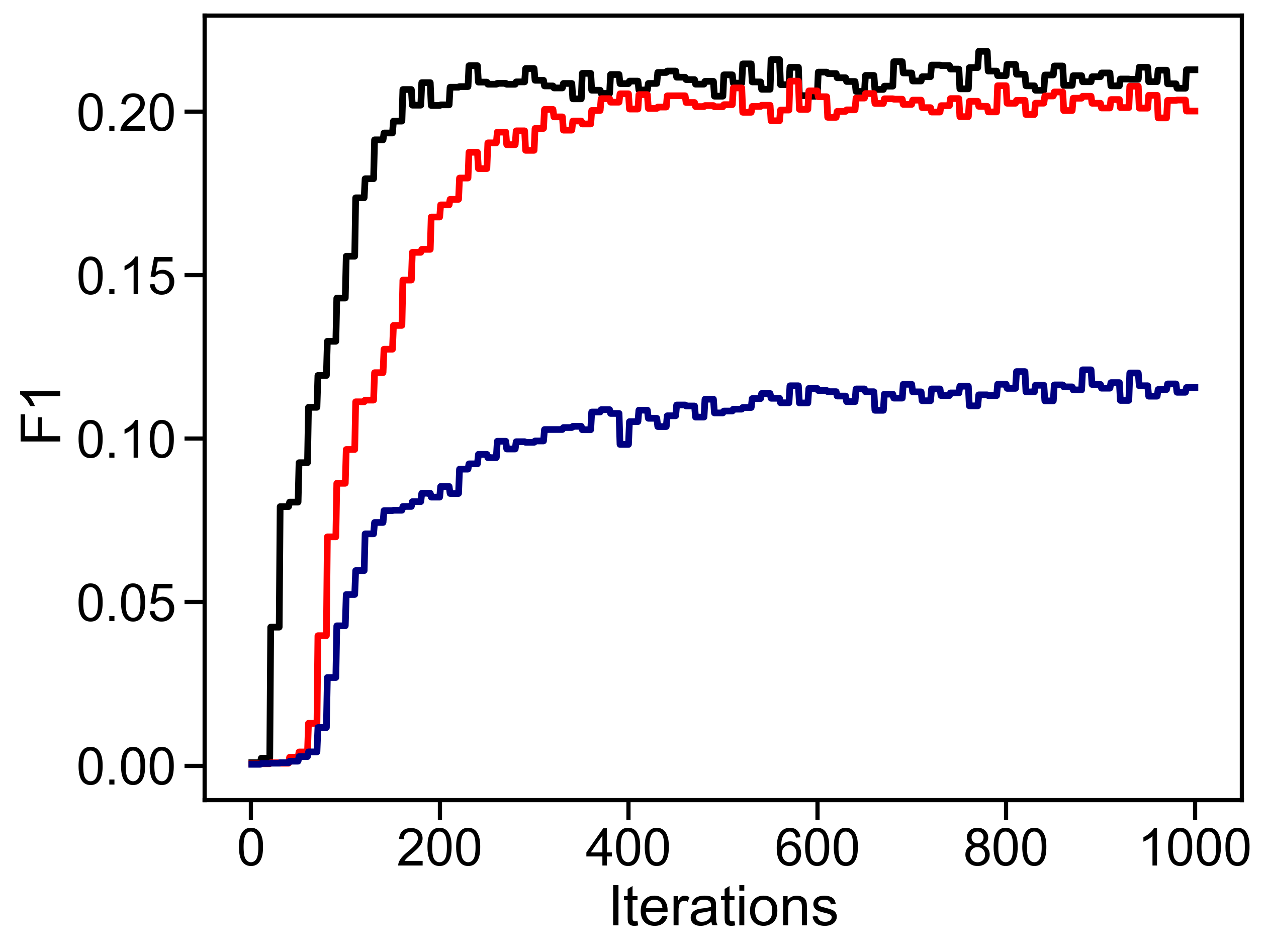}
	\includegraphics[width=0.24\textwidth]{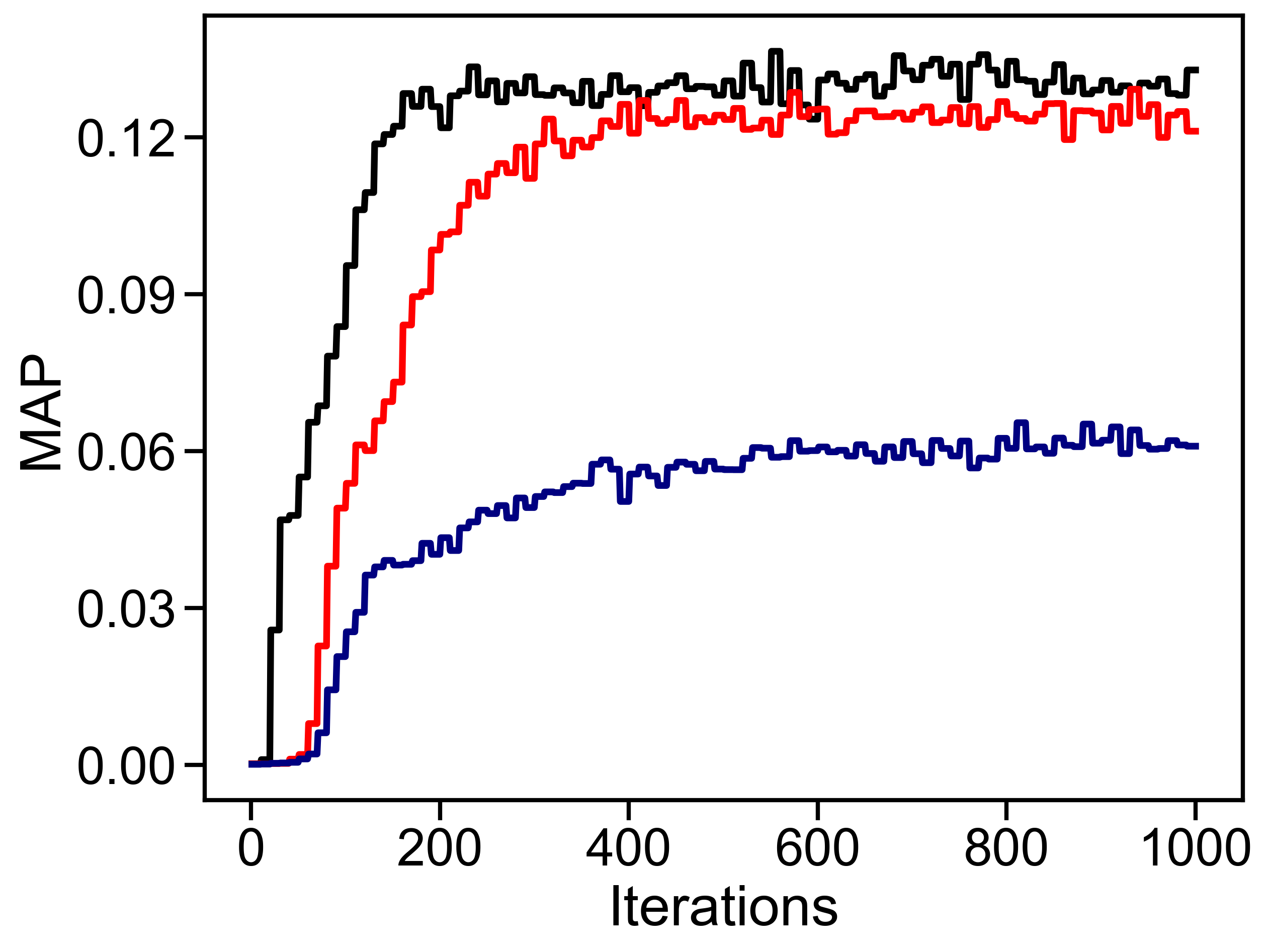}
	\begin{center}
		MIND
	\end{center}
	\includegraphics[width=0.24\textwidth]{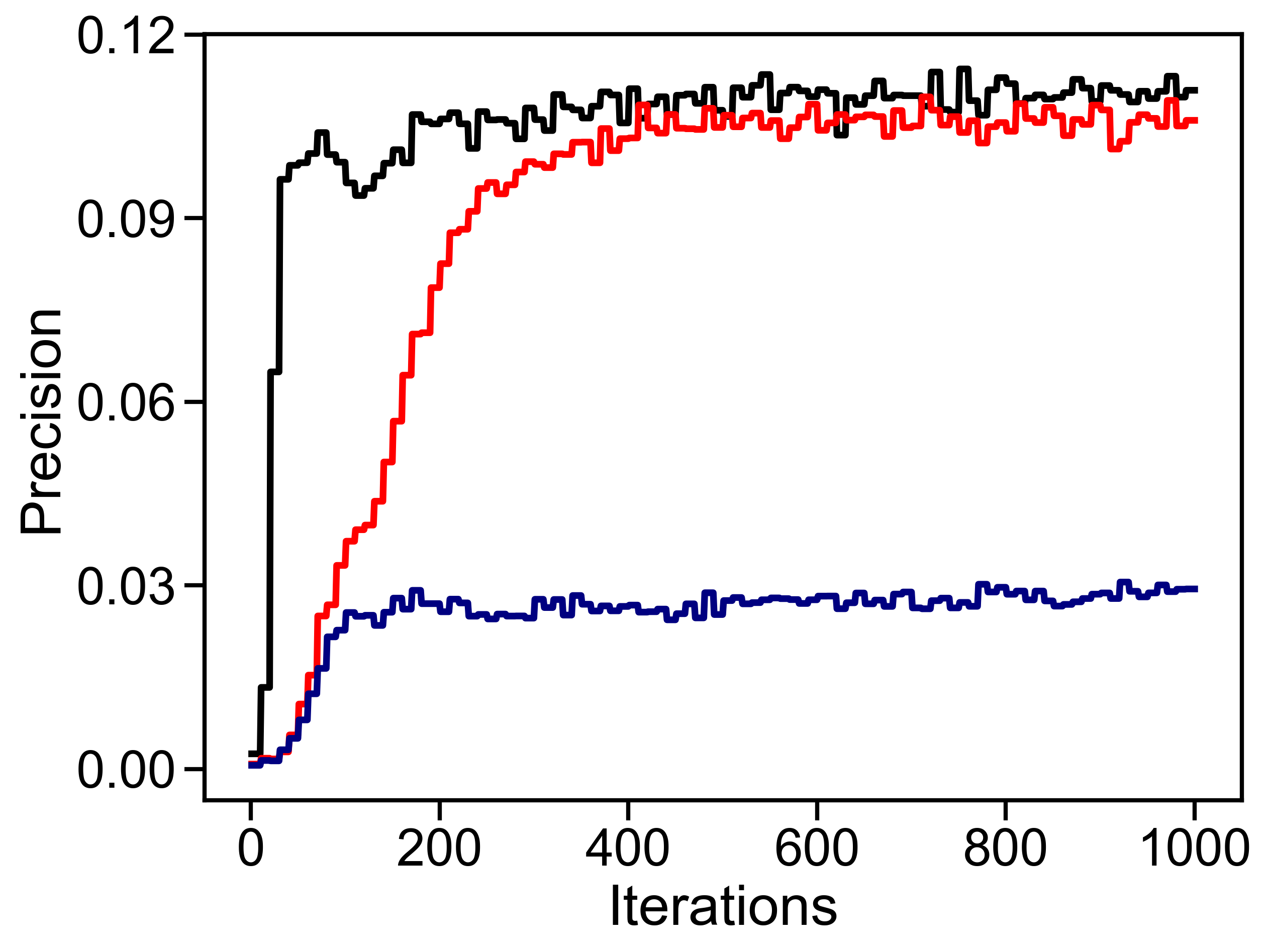}
	\includegraphics[width=0.24\textwidth]{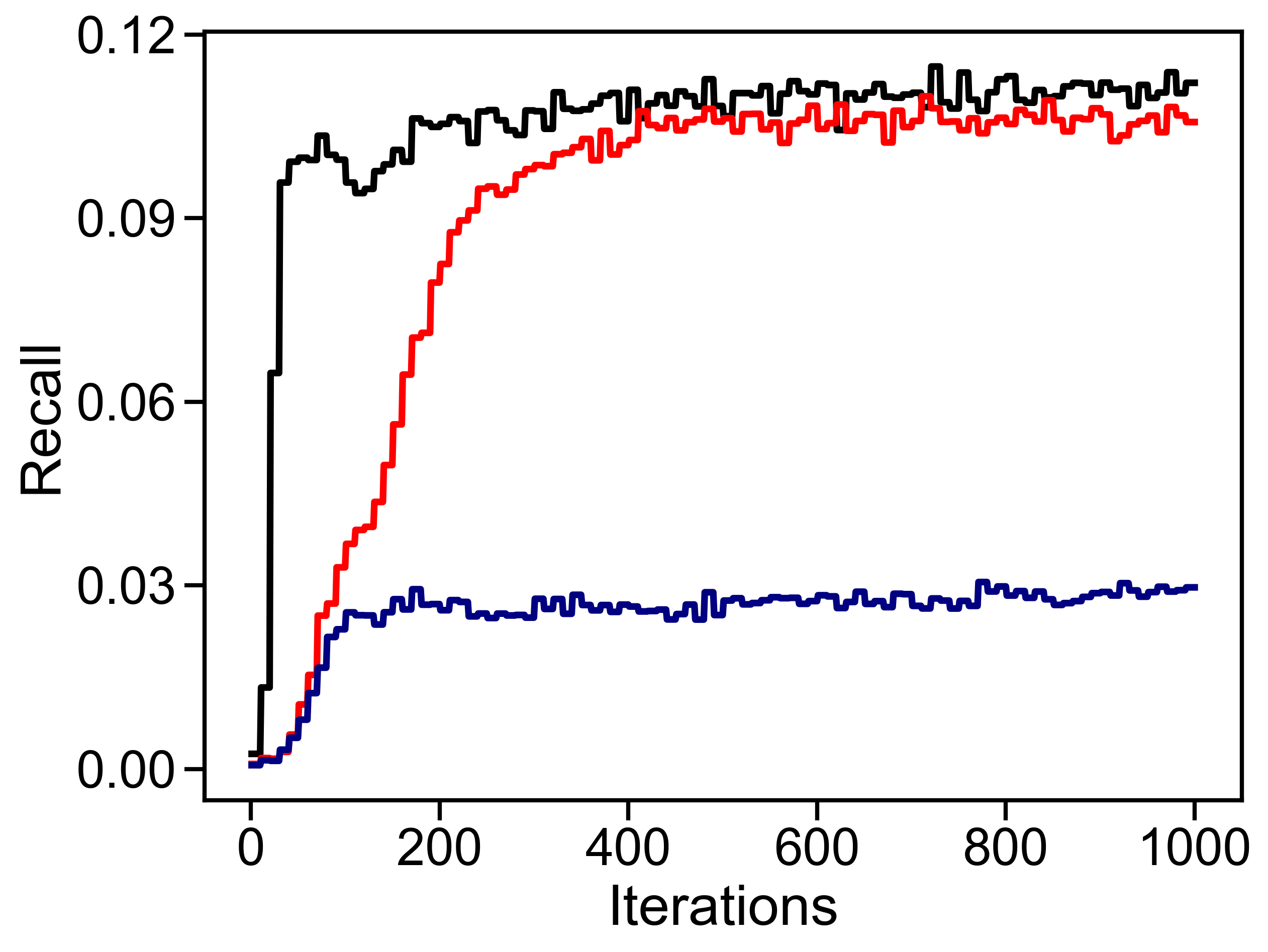}
	\includegraphics[width=0.24\textwidth]{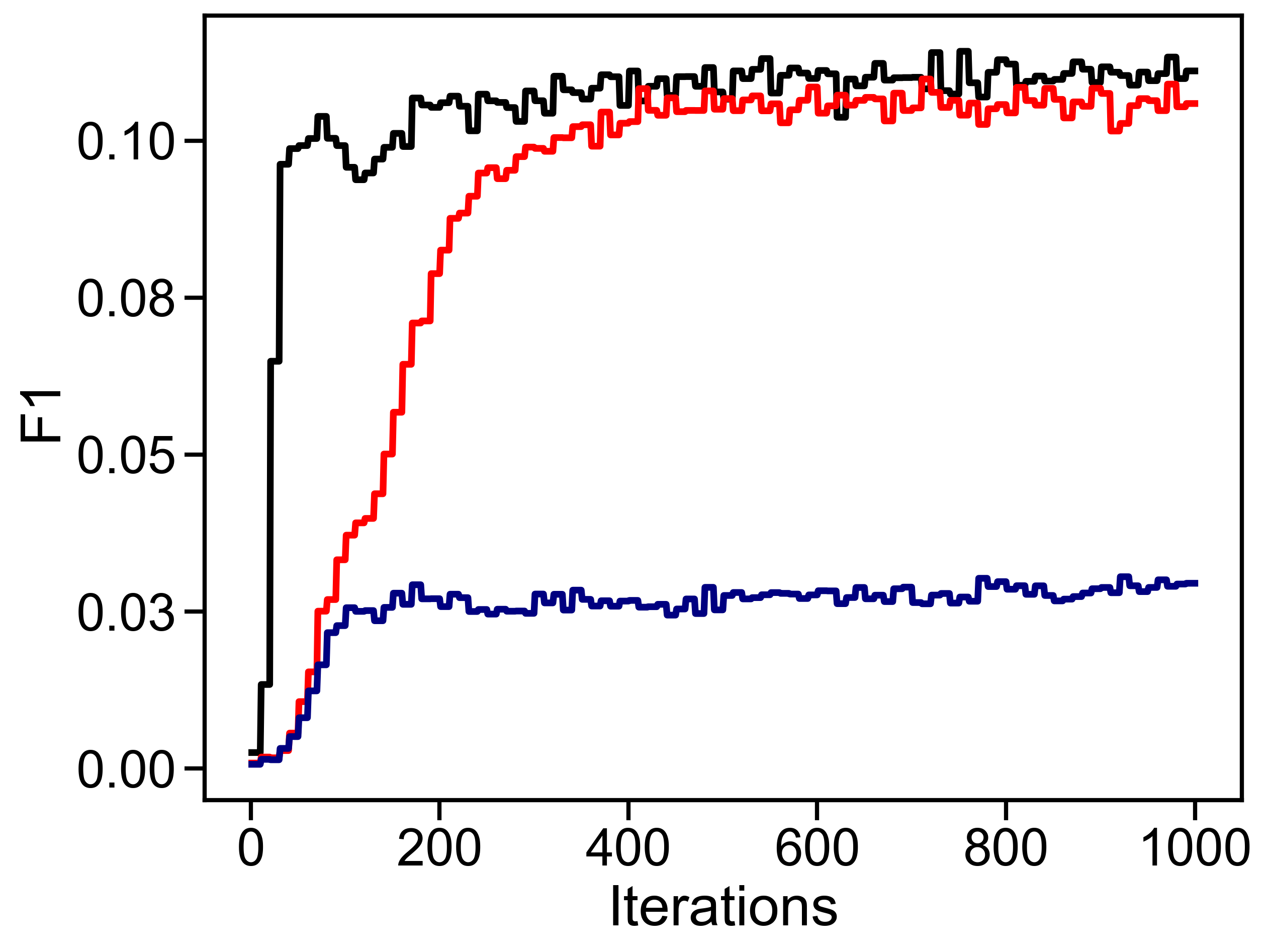}
	\includegraphics[width=0.24\textwidth]{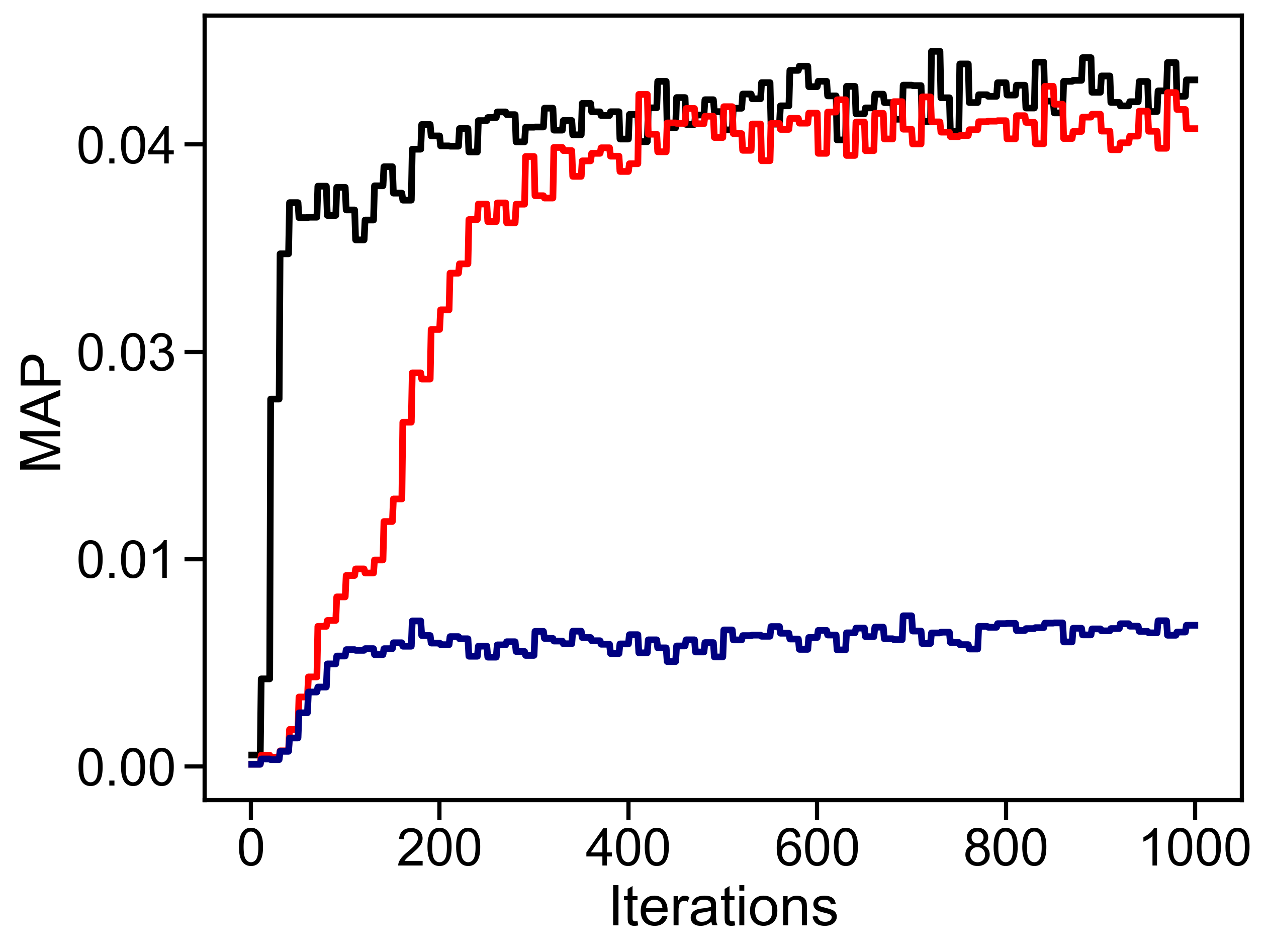}
	\includegraphics[width=0.5\textwidth]{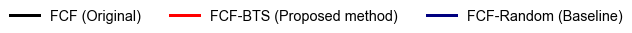}
	\caption{Convergence analysis of the proposed FCF-BTS method in a 90\% payload reduction scenario. X-axis shows the FL iterations. The Y-axis represents the performance metric values. For each FL iteration, we took the average of the previous ten metric values to account for the biases originating from the user’s unequal test set distributions. Each line denotes the average test set recommendation performance across 3 rounds of model rebuilds. In $\sim 400-450$ iterations, FCF-BTS converges to the best performance, which is close to FCF (Original) for the sparse dataset (Last-FM and MIND), while minimizing the payload by up to 90\%.}
	\label{fig:convergence}
\end{figure}

Next, we demonstrated that the proposed FCF-BTS method converges on the optimum and closely matches the solution that is achieved by FCF (original) for the sparse dataset (Last-FM and MIND). Figure~\ref{fig:convergence} shows that FCF (Original) reached the optimal solution between $\sim 200 - 250$ FL iterations in all three datasets. For the Last-FM and MIND datasets, we observed that the FCF-BTS method converges on the optimal solution between $\sim 400-450$, thus requiring $200$ additional iterations to get close to the upper-bound optimal solution as shown in Figure~\ref{fig:convergence}. This is typically expected in any form of optimization method that uses part of the whole model (fewer parameters) in each iteration. Most importantly, it validates the fact that FCF-BTS converges on the optimal solution while using only 10\% of the model payload, compared to the naive FCF-Random (baseline) method. In the Movielens dataset, we realized that FCF-BTS converges on the optimum in $\sim 200 - 250$ iterations similar to FCF (Original). However, the differences in performance are relatively large compared to the Last-FM and MIND datasets. Nevertheless, Figure~\ref{fig:convergence} illustrates the convergence stability of FCF-BTS across the three datasets up to 1,000 FL iterations similar to the FCF (Original) method’s convergence. In summary, our rigorous analysis confirms that the FCF-BTS solution closely matches the FCF (Original) method’s optimal solution for sparse datasets, although at a different rate. The results summarize that with a loss in the recommendation accuracy of  $\sim 4\% - 8\%$ (for highly sparse datasets) in comparison to the standard FCF method, FCF-BTS makes it possible to utilize a smaller payload (reduction up to 90\%) in FL model training. 

\section{Conclusion} \label{sec:conclusion}
In this study, we tackled the challenge of increasing payloads faced by FRS if deployed in a real-world situation. The requirement to move huge model payloads between the FL server and the user over several training rounds is neither practical nor feasible for a RS operating in production. We introduced an optimization method that addresses the payload challenge by selecting part (smaller payload) of the global model to be transmitted to all users. The selection process was guided by a bandit model optimizing a novel reward policy suitable for FRS. The proposed method was rigorously tested on three benchmark recommendation datasets and the empirical results demonstrate that our method consistently performed better compared to the simpler and naive optimization approaches. Our method achieved a 90\% reduction in payload with a minimal loss of recommendation performance from $\sim$4\% to 8\% in highly sparse datasets.  In addition, our method yielded a performance comparable to TopList with a 95\% payload reduction in two out of three datasets. The results establish that the bandit-based payload optimization can provide a similar quality of recommendation without increasing the computational cost for the user’s devices when participating in the FRS, particularly in production. 

In future work, we intend to extend the current research work in multiple directions. We have presented the payload optimization of the standard FCF to demonstrate a proof-of-concept. It would be interesting to investigate whether similar results will be achieved in the context of larger datasets and far more recent and advanced FRS methods~\cite{qi2020privacy,10.1007/978-3-030-67661-2_20}. In this study, we empirically validated the usefulness of the proposed optimization method. A key next step would be to study the theoretical properties reflecting upon the convergence guarantees and regret bounds for the novel reward function.  

\paragraph{Acknowledgement:}
This work was supported by Helsinki Research Center, Europe Cloud Service Competence Center, Huawei Technologies Oy (Finland) Co. Ltd.

\bibliographystyle{unsrtnat}
\bibliography{references}

\begin{thebibliography}{45}
\providecommand{\natexlab}[1]{#1}
\providecommand{\url}[1]{\texttt{#1}}
\expandafter\ifx\csname urlstyle\endcsname\relax
  \providecommand{\doi}[1]{doi: #1}\else
  \providecommand{\doi}{doi: \begingroup \urlstyle{rm}\Url}\fi

\bibitem[McMahan et~al.(2017)McMahan, Moore, Ramage, Hampson, and
  y~Arcas]{mcmahan2017communication}
Brendan McMahan, Eider Moore, Daniel Ramage, Seth Hampson, and Blaise~Aguera
  y~Arcas.
\newblock Communication-efficient learning of deep networks from decentralized
  data.
\newblock In \emph{Artificial Intelligence and Statistics}, pages 1273--1282,
  2017.

\bibitem[Ammad-Ud-Din et~al.(2019)Ammad-Ud-Din, Ivannikova, Khan, Oyomno, Fu,
  Tan, and Flanagan]{ammad2019federated}
Muhammad Ammad-Ud-Din, Elena Ivannikova, Suleiman~A Khan, Were Oyomno, Qiang
  Fu, Kuan~Eeik Tan, and Adrian Flanagan.
\newblock Federated collaborative filtering for privacy-preserving personalized
  recommendation system.
\newblock \emph{arXiv preprint arXiv:1901.09888}, 2019.

\bibitem[Chai et~al.(2020)Chai, Wang, Chen, and Yang]{chai2019secure}
Di~Chai, Leye Wang, Kai Chen, and Qiang Yang.
\newblock Secure federated matrix factorization.
\newblock \emph{IEEE Intelligent Systems}, pages 1--1, 08 2020.
\newblock \doi{10.1109/MIS.2020.3014880}.

\bibitem[Dolui et~al.(2019)Dolui, Gyllensten, Lowet, Michiels, Hallez, and
  Hughes]{dolui2019poster}
Koustabh Dolui, Illapha~Cuba Gyllensten, Dietwig Lowet, Sam Michiels, Hans
  Hallez, and Danny Hughes.
\newblock Poster: Towards privacy-preserving mobile applications with federated
  learning--the case of matrix factorization.
\newblock In \emph{The 17th Annual International Conference on Mobile Systems,
  Applications, and Services}, page 624–625, 2019.

\bibitem[Hu et~al.(2008)Hu, Koren, and Volinsky]{Hu2008}
Yifan Hu, Yehuda Koren, and Chris Volinsky.
\newblock Collaborative filtering for implicit feedback datasets.
\newblock In \emph{Proceedings of the 2008 Eighth IEEE International Conference
  on Data Mining}, ICDM '08, pages 263--272, Washington, DC, USA, 2008. IEEE
  Computer Society.
\newblock ISBN 978-0-7695-3502-9.

\bibitem[Koren et~al.(2009)Koren, Bell, and Volinsky]{koren2009matrix}
Yehuda Koren, Robert~M. Bell, and Chris Volinsky.
\newblock Matrix factorization techniques for recommender systems.
\newblock \emph{{IEEE} Computer}, 42\penalty0 (8):\penalty0 30--37, 2009.
\newblock \doi{10.1109/MC.2009.263}.
\newblock URL \url{https://doi.org/10.1109/MC.2009.263}.

\bibitem[Li et~al.(2020)Li, Sahu, Talwalkar, and Smith]{litian2019federated}
Tian Li, Anit~Kumar Sahu, Ameet Talwalkar, and Virginia Smith.
\newblock Federated learning: Challenges, methods, and future directions.
\newblock \emph{IEEE Signal Processing Magazine}, 37\penalty0 (3):\penalty0
  50--60, 2020.
\newblock \doi{10.1109/MSP.2020.2975749}.

\bibitem[Li et~al.(2019)Li, Wen, and He]{li2019federated}
Qinbin Li, Zeyi Wen, and Bingsheng He.
\newblock Federated learning systems: Vision, hype and reality for data privacy
  and protection.
\newblock \emph{arXiv preprint arXiv:1907.09693}, 2019.

\bibitem[Flanagan et~al.(2021)Flanagan, Oyomno, Grigorievskiy, Tan, Khan, and
  Ammad-Ud-Din]{10.1007/978-3-030-67661-2_20}
Adrian Flanagan, Were Oyomno, Alexander Grigorievskiy, Kuan~E. Tan, Suleiman~A.
  Khan, and Muhammad Ammad-Ud-Din.
\newblock Federated multi-view matrix factorization for personalized
  recommendations.
\newblock In \emph{Machine Learning and Knowledge Discovery in Databases},
  pages 324--347, Cham, 2021. Springer International Publishing.
\newblock ISBN 978-3-030-67661-2.

\bibitem[Qi et~al.(2020)Qi, Wu, Wu, Huang, and Xie]{qi2020privacy}
Tao Qi, Fangzhao Wu, Chuhan Wu, Yongfeng Huang, and Xing Xie.
\newblock Privacy-preserving news recommendation model training via federated
  learning.
\newblock \emph{arXiv preprint arXiv:2003.09592}, 2020.

\bibitem[Kingma and Ba(2015)]{kingma2015adam}
Diederik~P Kingma and Jimmy~Lei Ba.
\newblock Adam: {A} method for stochastic optimization.
\newblock In \emph{International Conference on Learning Representations}, 2015.

\bibitem[Thompson(1933)]{thompson1933likelihood}
William~R Thompson.
\newblock On the likelihood that one unknown probability exceeds another in
  view of the evidence of two samples.
\newblock \emph{Biometrika}, 25\penalty0 (3/4):\penalty0 285--294, 1933.

\bibitem[Thompson(1935)]{thompson1935theory}
William~R Thompson.
\newblock On the theory of apportionment.
\newblock \emph{American Journal of Mathematics}, 57\penalty0 (2):\penalty0
  450--456, 1935.

\bibitem[Chapelle and Li(2011)]{chapelle2011empirical}
Olivier Chapelle and Lihong Li.
\newblock An empirical evaluation of thompson sampling.
\newblock \emph{Advances in neural information processing systems},
  24:\penalty0 2249--2257, 2011.

\bibitem[Scott(2010)]{scott2010modern}
Steven~L Scott.
\newblock A modern bayesian look at the multi-armed bandit.
\newblock \emph{Applied Stochastic Models in Business and Industry},
  26\penalty0 (6):\penalty0 639--658, 2010.

\bibitem[Kawale et~al.(2015)Kawale, Bui, Kveton, Tran-Thanh, and
  Chawla]{kawale2015efficient}
Jaya Kawale, Hung~H Bui, Branislav Kveton, Long Tran-Thanh, and Sanjay Chawla.
\newblock Efficient thompson sampling for online matrix-factorization
  recommendation.
\newblock In \emph{Advances in neural information processing systems}, pages
  1297--1305, 2015.

\bibitem[Gelman et~al.(2013)Gelman, Carlin, Stern, Dunson, Vehtari, and
  Rubin]{gelman2013bayesian}
Andrew Gelman, John~B Carlin, Hal~S Stern, David~B Dunson, Aki Vehtari, and
  Donald~B Rubin.
\newblock \emph{Bayesian data analysis}.
\newblock CRC press, 2013.

\bibitem[Fink(1997)]{fink1997compendium}
Daniel Fink.
\newblock A compendium of conjugate priors.
\newblock \emph{See http://www. people. cornell. edu/pages/df36/CONJINTRnew\%
  20TEX. pdf}, 46, 1997.

\bibitem[Streeter and Golovin(2008)]{streeter2008online}
Matthew Streeter and Daniel Golovin.
\newblock An online algorithm for maximizing submodular functions.
\newblock In \emph{Proceedings of the 21st International Conference on Neural
  Information Processing Systems}, pages 1577--1584, 2008.

\bibitem[Radlinski et~al.(2008)Radlinski, Kleinberg, and
  Joachims]{radlinski2008learning}
Filip Radlinski, Robert Kleinberg, and Thorsten Joachims.
\newblock Learning diverse rankings with multi-armed bandits.
\newblock In \emph{Proceedings of the 25th international conference on Machine
  learning}, pages 784--791, 2008.

\bibitem[Uchiya et~al.(2010)Uchiya, Nakamura, and Kudo]{uchiya2010algorithms}
Taishi Uchiya, Atsuyoshi Nakamura, and Mineichi Kudo.
\newblock Algorithms for adversarial bandit problems with multiple plays.
\newblock In \emph{International Conference on Algorithmic Learning Theory},
  pages 375--389. Springer, 2010.

\bibitem[Lou{\"e}dec et~al.(2015)Lou{\"e}dec, Chevalier, Mothe, Garivier, and
  Gerchinovitz]{louedec2015multiple}
Jonathan Lou{\"e}dec, Max Chevalier, Josiane Mothe, Aur{\'e}lien Garivier, and
  S{\'e}bastien Gerchinovitz.
\newblock A multiple-play bandit algorithm applied to recommender systems.
\newblock In \emph{The Twenty-Eighth International Flairs Conference}, 2015.

\bibitem[Gopalan et~al.(2014)Gopalan, Mannor, and Mansour]{gopalan2014thompson}
Aditya Gopalan, Shie Mannor, and Yishay Mansour.
\newblock Thompson sampling for complex online problems.
\newblock In \emph{International Conference on Machine Learning}, pages
  100--108. PMLR, 2014.

\bibitem[Brod{\'e}n et~al.(2018)Brod{\'e}n, Hammar, Nilsson, and
  Paraschakis]{broden2018ensemble}
Bj{\"o}rn Brod{\'e}n, Mikael Hammar, Bengt~J Nilsson, and Dimitris Paraschakis.
\newblock Ensemble recommendations via thompson sampling: an experimental study
  within e-commerce.
\newblock In \emph{23rd international conference on intelligent user
  interfaces}, pages 19--29, 2018.

\bibitem[Agrawal and Goyal(2013)]{agrawal2013further}
Shipra Agrawal and Navin Goyal.
\newblock Further optimal regret bounds for thompson sampling.
\newblock In \emph{Artificial intelligence and statistics}, pages 99--107.
  PMLR, 2013.

\bibitem[Russo and Van~Roy(2016)]{russo2016information}
Daniel Russo and Benjamin Van~Roy.
\newblock An information-theoretic analysis of thompson sampling.
\newblock \emph{The Journal of Machine Learning Research}, 17\penalty0
  (1):\penalty0 2442--2471, 2016.

\bibitem[Dong and Roy(2018)]{dong2018information}
Shi Dong and Benjamin~Van Roy.
\newblock An information-theoretic analysis for thompson sampling with many
  actions.
\newblock In \emph{Proceedings of the 32nd International Conference on Neural
  Information Processing Systems}, pages 4161--4169, 2018.

\bibitem[Konecn{\'{y}} et~al.(2016)Konecn{\'{y}}, McMahan, Yu, Richt{\'{a}}rik,
  Suresh, and Bacon]{DBLP:journals/corr/KonecnyMYRSB16}
Jakub Konecn{\'{y}}, H.~Brendan McMahan, Felix~X. Yu, Peter Richt{\'{a}}rik,
  Ananda~Theertha Suresh, and Dave Bacon.
\newblock Federated learning: Strategies for improving communication
  efficiency.
\newblock \emph{CoRR}, abs/1610.05492, 2016.
\newblock URL \url{http://arxiv.org/abs/1610.05492}.

\bibitem[Han et~al.(2020)Han, Wang, and Leung]{han2020adaptive}
Pengchao Han, Shiqiang Wang, and Kin~K Leung.
\newblock Adaptive gradient sparsification for efficient federated learning: An
  online learning approach.
\newblock \emph{arXiv preprint arXiv:2001.04756}, 2020.

\bibitem[Sattler et~al.(2019)Sattler, Wiedemann, M{\"u}ller, and
  Samek]{sattler2019robust}
Felix Sattler, Simon Wiedemann, Klaus-Robert M{\"u}ller, and Wojciech Samek.
\newblock Robust and communication-efficient federated learning from non-iid
  data.
\newblock \emph{IEEE transactions on neural networks and learning systems},
  31\penalty0 (9):\penalty0 3400--3413, 2019.

\bibitem[Jeong et~al.(2018)Jeong, Oh, Kim, Park, Bennis, and
  Kim]{jeong2018federated}
E~Jeong, S~Oh, H~Kim, J~Park, M~Bennis, and SL~Kim.
\newblock Federated distillation and augmentation under non-iid private data.
\newblock \emph{NIPS Wksp. MLPCD}, 2018.

\bibitem[He et~al.(2020)He, Annavaram, and Avestimehr]{he2020group}
Chaoyang He, Murali Annavaram, and Salman Avestimehr.
\newblock Group knowledge transfer: Federated learning of large cnns at the
  edge.
\newblock \emph{Advances in Neural Information Processing Systems}, 33, 2020.

\bibitem[Caldas et~al.(2018)Caldas, Kone{\v{c}}ny, McMahan, and
  Talwalkar]{caldas2018expanding}
Sebastian Caldas, Jakub Kone{\v{c}}ny, H~Brendan McMahan, and Ameet Talwalkar.
\newblock Expanding the reach of federated learning by reducing client resource
  requirements.
\newblock \emph{arXiv preprint arXiv:1812.07210}, 2018.

\bibitem[Saputra et~al.(2019)Saputra, Hoang, Nguyen, Dutkiewicz, Mueck, and
  Srikanteswara]{saputra2019energy}
Yuris~Mulya Saputra, Dinh~Thai Hoang, Diep~N Nguyen, Eryk Dutkiewicz,
  Markus~Dominik Mueck, and Srikathyayani Srikanteswara.
\newblock Energy demand prediction with federated learning for electric vehicle
  networks.
\newblock In \emph{2019 IEEE Global Communications Conference (GLOBECOM)},
  pages 1--6. IEEE, 2019.

\bibitem[Dai et~al.(2020)Dai, Low, and Jaillet]{NEURIPS2020_6dfe08ed}
Zhongxiang Dai, Bryan Kian~Hsiang Low, and Patrick Jaillet.
\newblock Federated bayesian optimization via thompson sampling.
\newblock In H.~Larochelle, M.~Ranzato, R.~Hadsell, M.~F. Balcan, and H.~Lin,
  editors, \emph{Advances in Neural Information Processing Systems}, volume~33,
  pages 9687--9699. Curran Associates, Inc., 2020.
\newblock URL
  \url{https://proceedings.neurips.cc/paper/2020/file/6dfe08eda761bd321f8a9b239f6f4ec3-Paper.pdf}.

\bibitem[Dubey and Pentland(2020)]{NEURIPS2020_4311359e}
Abhimanyu Dubey and Alex \textasciigrave~Sandy\textquotesingle Pentland.
\newblock Differentially-private federated linear bandits.
\newblock In H.~Larochelle, M.~Ranzato, R.~Hadsell, M.~F. Balcan, and H.~Lin,
  editors, \emph{Advances in Neural Information Processing Systems}, volume~33,
  pages 6003--6014. Curran Associates, Inc., 2020.
\newblock URL
  \url{https://proceedings.neurips.cc/paper/2020/file/4311359ed4969e8401880e3c1836fbe1-Paper.pdf}.

\bibitem[Zhou et~al.(2019)Zhou, Wang, Guo, Gong, and Zheng]{zhou2019privacy}
Pan Zhou, Kehao Wang, Linke Guo, Shimin Gong, and Bolong Zheng.
\newblock A privacy-preserving distributed contextual federated online learning
  framework with big data support in social recommender systems.
\newblock \emph{IEEE Transactions on Knowledge and Data Engineering}, 2019.

\bibitem[Tan et~al.(2020)Tan, Liu, Zheng, and Yang]{tan2020federated}
Ben Tan, Bo~Liu, Vincent Zheng, and Qiang Yang.
\newblock A federated recommender system for online services.
\newblock In \emph{Fourteenth ACM Conference on Recommender Systems}, pages
  579--581, 2020.

\bibitem[Chen et~al.(2018)Chen, Dong, Li, and He]{chen2018federated}
Fei Chen, Zhenhua Dong, Zhenguo Li, and Xiuqiang He.
\newblock Federated meta-learning for recommendation.
\newblock \emph{arXiv preprint arXiv:1802.07876}, 2018.

\bibitem[Muhammad et~al.(2020)Muhammad, Wang, O'Reilly-Morgan, Tragos, Smyth,
  Hurley, Geraci, and Lawlor]{muhammad2020fedfast}
Khalil Muhammad, Qinqin Wang, Diarmuid O'Reilly-Morgan, Elias Tragos, Barry
  Smyth, Neil Hurley, James Geraci, and Aonghus Lawlor.
\newblock Fedfast: Going beyond average for faster training of federated
  recommender systems.
\newblock In \emph{Proceedings of the 26th ACM SIGKDD International Conference
  on Knowledge Discovery \& Data Mining}, pages 1234--1242, 2020.

\bibitem[Qin and Liu(2020)]{qin2020novel}
Jiangcheng Qin and Baisong Liu.
\newblock A novel privacy-preserved recommender system framework based on
  federated learning.
\newblock \emph{arXiv preprint arXiv:2011.05614}, 2020.

\bibitem[Harper and Konstan(2015)]{harper2015movielens}
F~Maxwell Harper and Joseph~A Konstan.
\newblock The movielens datasets: History and context.
\newblock \emph{Acm transactions on interactive intelligent systems (tiis)},
  5\penalty0 (4):\penalty0 1--19, 2015.

\bibitem[Cantador et~al.(2011)Cantador, Brusilovsky, and
  Kuflik]{cantador2011second}
Iv{\'a}n Cantador, Peter Brusilovsky, and Tsvi Kuflik.
\newblock Second workshop on information heterogeneity and fusion in
  recommender systems (hetrec2011).
\newblock In \emph{Proceedings of the fifth ACM conference on Recommender
  systems}, pages 387--388, 2011.

\bibitem[Wu et~al.(2020)Wu, Qiao, Chen, Wu, Qi, Lian, Liu, Xie, Gao, Wu,
  et~al.]{wu2020mind}
Fangzhao Wu, Ying Qiao, Jiun-Hung Chen, Chuhan Wu, Tao Qi, Jianxun Lian,
  Danyang Liu, Xing Xie, Jianfeng Gao, Winnie Wu, et~al.
\newblock Mind: A large-scale dataset for news recommendation.
\newblock In \emph{Proceedings of the 58th Annual Meeting of the Association
  for Computational Linguistics}, pages 3597--3606, 2020.

\bibitem[Bobadilla et~al.(2013)Bobadilla, Ortega, Hernando, and
  Guti{\'e}rrez]{bobadilla2013recommender}
Jes{\'u}s Bobadilla, Fernando Ortega, Antonio Hernando, and Abraham
  Guti{\'e}rrez.
\newblock Recommender systems survey.
\newblock \emph{Knowledge-based systems}, 46:\penalty0 109--132, 2013.

\end{thebibliography}
\end{document}